\documentclass[a4paper]{article}

\usepackage[utf8]{inputenc}

\usepackage{authblk} 
\usepackage{geometry} 
\usepackage{hyperref} 
\usepackage{amsmath} 
\usepackage{graphicx}
\usepackage{natbib} 

\usepackage{fancyhdr} 
\fancypagestyle{plain}{
\fancyhf{}
\fancyhead[c]{This manuscript has been accepted for publication in \emph{Bioinformatics} published by Oxford University Press: \url{https://doi.org/10.1093/bioinformatics/btz199} (open access)}
}

\begin{document}

\title{Simulation assisted machine learning}
\author[1, 2]{Timo M. Deist\footnote{These authors contributed equally to this work}\textsuperscript{,}}
\author[1]{Andrew Patti\textsuperscript{$\ast$,}}
\author[1]{Zhaoqi Wang}
\author[1]{David Krane}
\author[1]{Taylor Sorenson}
\author[1]{David Craft}

\affil[1]{Department of Radiation Oncology, Massachusetts General Hospital, Harvard Medical School}
\affil[2]{The D-Lab: Decision Support for Precision Medicine, GROW - School for Oncology and Developmental Biology, Maastricht University Medical Centre}

\maketitle
\abstract{\noindent \textbf{Motivation:} In a predictive modeling setting, if sufficient details of the system behavior are known, one can build and use a simulation for making predictions. When sufficient system details are not known, one typically turns to machine learning, which builds a black-box model of the system using a large dataset of input sample features and outputs.
We consider a setting which is between these two extremes: some details of the system mechanics are known but not enough for creating simulations that can be used to make high quality predictions. In this context we propose using approximate simulations to build a kernel for use in kernelized machine learning methods, such as support vector machines. The results of multiple simulations (under various uncertainty scenarios) are used to compute similarity measures between every pair of samples: sample pairs are given a high similarity score if they behave similarly under a wide range of simulation parameters. These similarity values, rather than the original high dimensional feature data, are used to build the kernel. \\
\textbf{Results:} We demonstrate and explore the simulation based kernel (SimKern) concept using four synthetic complex systems--three biologically inspired models and one network flow optimization model. We show that, when the number of training samples is small compared to the number of features, the SimKern approach dominates over no-prior-knowledge methods. This approach should be applicable in all disciplines where predictive models are sought and informative yet approximate simulations are available.\\
\textbf{Availability:} The Python SimKern software, the demonstration models (in MATLAB, R), and the datasets are available at \url{https://github.com/davidcraft/SimKern}.\\
\textbf{Contact:} \href{dcraft@broadinstitute.org}{dcraft@broadinstitute.org}\\
\textbf{Supplementary information:} Supplementary data are available.}

\section{Introduction and motivation}

There are two general approaches to computationally predicting the behavior of complex systems, simulation and machine learning (ML). Simulation is the preferred method if the dynamics of the system being studied are known in sufficient detail that one can simulate its behavior with high fidelity and map the system behavior to the output to be predicted. ML is valuable when the system defies accurate simulation but enough data exists to train a general black-box machine learner, which could be anything from a linear regression or classification model to a neural network. In this work, we propose a technique to combine simulation and ML in order to leverage the best aspects of both and produce a system that is superior to either technique alone.

Our motivation is personalized medicine: how do we assign the right drug or drug combination to cancer patients? Across cultures and history, physicians prescribe medicines and interventions based on how the patient is predicted to respond. Currently these choices are made based on established patient-classification protocols, physician judgment, clinical trial eligibility, and occasionally limited genomic profiling of the patient. All of these approaches, in one way or another, attempt to partition patients into groups based on some notion of similarity.

Genomics is especially relevant for computing the similarity between two cancer patients since cancer is associated with alterations to the DNA, which in turn causes the dysregulation of cellular behavior \citep{balmain}. Bioinformatic analysis has revealed that there is heterogeneity both within a patient tumor and across tumors; no two tumors are the same genomically \citep{hetero,hetero2}.  Although in a small fraction of cases specific genetic conditions are used to guide therapy choices, for example breast (commonly amplified gene: HER2), melanoma (BRAF mutation), lung (EML4-ALK fusion), and head-and-neck (HPV status for radiation dose de-escalation \citep{hpv}), there remains a large variability in patient responses to these and other treatments, likely due to the fact that patients will usually have tens or hundreds of mutations and gene copy number variations, chromosomal structural rearrangements, not to mention a distinct germline genetic state \citep{gpcr}, human leukocyte antigen type \citep{hlaImmuno}, tumor epigenetic DNA modifications, microbiome, and comorbidity set. Even amidst this heterogeneity, the notion of patient similarity--although currently not deeply understood due to the complexities of cancer biology--is appealing both conceptually and for its value in the ML setting.

Simulating a drug is a task that far exceeds our current scientific capacity: it enters the patient, either intravenously or orally, and winds its way to the cancer cells, where it either influences the cancer cell via receptors on the cell membrane or penetrates into the cell and affects signaling pathways, cell metabolism, DNA repair, apoptosis, or some combination of these and other modules. Nevertheless, a vast amount of knowledge of cellular processes, residing in molecular biology textbooks and millions of scientific papers, has been accrued over the past century and it seems worthwhile to attempt to use that information, if unclear how. Most machine learning research efforts in the personalized medicine realm take a pure data approach. Given the complexity of patient biology and cancer, this approach will require vast amounts of high quality patient data that is suitably standardized for algorithmic processing. 

With this drug sensitivity prediction problem as our backdrop, we develop a method to combine approximate simulations with ML and demonstrate using {\em in silico} experiments that a judicious combination can yield better predictions than either technique alone. The basic idea is a division of labor: coarse and approximate simulations are used to compute similarity measures, and these similarity measures are then used by the ML algorithm to build a predictive model, called SimKern ML (Figure~1). At this point in time, although vast details of cellular biology are known, we are not in a position to simulate with any fidelity complete cellular or {\em in vivo} cancerous processes. However, herein we present demonstrations that one could combine simulation results into machine learning and improve the overall predictive capability, a technique which may play a role in future drug recommendation systems.

\begin{figure}
\centering
\includegraphics[trim=0 0 0 0,clip,width=16.7cm]{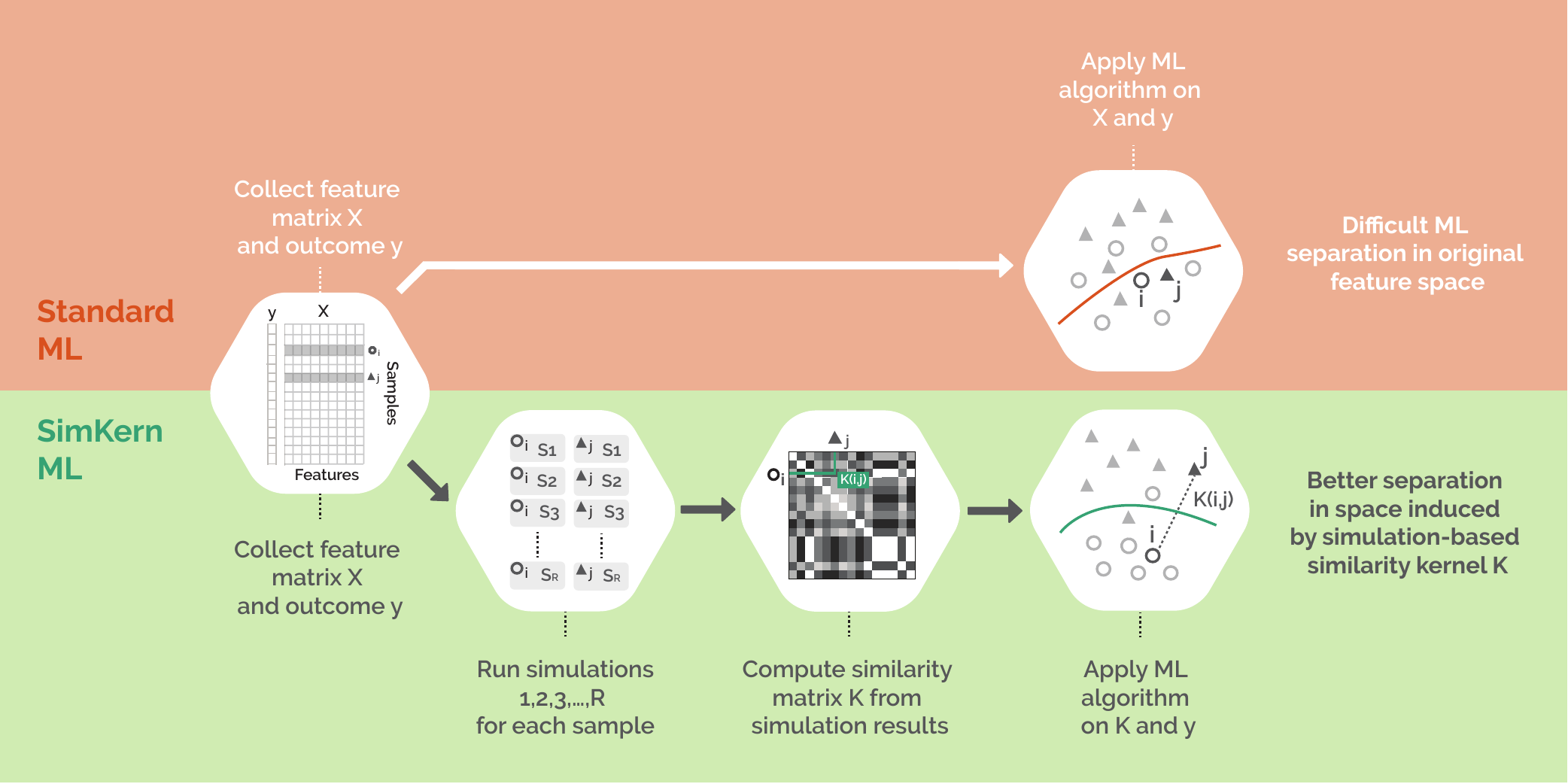}
\caption{Workflow comparison of Standard machine learning (ML) and SimKern ML. The feature matrix $X$ and outcome data $y$ are given (in this paper, we generate such ``ground truth'' datasets by simulating complex systems, a step which is not shown in this figure). Traditional feature-based ML is depicted in the upper orange part. SimKern, the simulation-based method, pre-processes the dataset by sending each sample through a number of approximate simulations. Each sample pair is given a similarity score based on how closely they behave under the various simulations (See Figure 2 and Section \ref{simkernGen} for more details). This information is stored in a kernel matrix $K$, where $K(i,j)$ measures the similarity between samples $i$ and $j$. Note that $K(i,i)=1$ and $0 \le K(i,j) \le 1$. Useful SimKern simulations yield a kernel $K$ that improves the downstream machine learning performance.}\label{schematic}
\end{figure}

\section{Materials and methods}
Our method is centered on kernelized ML. Rather than feature vectors (a list of attributes for each sample), kernelized learning requires only a similarity score between pairs of samples. For training, one needs the outcome of each training sample and a measurement of the similarity between all pairs of training samples. For predicting the outcome of a new sample, one needs to provide the similarity of that sample to each training sample. It is well known in ML that good similarity measures, which come from expert domain knowledge, result in better ML performance \citep{kernelBook}. We assume that we can formulate a simulation of each sample's behavior based on its known individual characteristics (i.e. features). We also assume that we do not know exactly how to simulate the systems, so rather than a single simulation we have a family (possibly parametrized by real numbers, and thus infinite) of plausible simulations. Two samples are given a high similarity score if they behave similarly across a wide range of simulations. 

We begin with a brief description of the four models we use to demonstrate and analyze the performance of SimKern. By describing these models, the reader has in mind a more concrete context with which to frame the SimKern development.

\subsection{Brief model descriptions}

We investigate four models: radiation impact on cells, flowering time in plants, a Boolean cancer model, and a network flow optimization problem. Full details and model implementation notes are given in the Supplementary information. 

For each model we begin by generating a dataset of $N$ samples, each sample $i$ is described by a feature vector $x_{i}$ of length $p$ and a response $y_{i}$, using the ground truth simulation (see Figure~S1). This produces an $N\times p$ feature matrix $X$ and a response vector $y$ of length $N$. This ground truth simulation (referred to as SIM0 in the code repository) is not part of our kernelized learning method, but the datasets created are needed to demonstrate the simulation-based kernel ML method. This ground truth simulation step is further described in the Supplementary information. In an actual application of SimKern, this artificial data creation step would not be used.

The {\bf radiation cancer cell death model} is a set of ordinary differential equations (ODEs) which represents a simplified view of the biochemical processes that happen after a cancer cell is hit by radiation. The core of the model involves the DNA damage response regulated by the phosphorylation of ATM and subsequent p53 tetramerization \citep{elias}. We have added cell cycle arrest terms, a DNA repair process, and apoptosis modules in order to capture the idea that cellular response to DNA damage involves the combined dynamics of these various processes. The model, which is depicted as a network graph, is displayed in Supplementary Figure S3, and consists of 34 ODEs. The rate parameters were not tuned to realistic values (except for the ones from the original p53 core network, where we used the values provided by the authors \citep{elias}). Instead, values were manually chosen such that the family of samples created had representatives in each of the four output classes: apoptosis, repaired and cycling, mitotic catastrophe, and quiescence. A population of distinct cell types is formed by varying 33 of the ODE rate constants and the mutation status of six genes (ARF, BAX, SIAH, Reprimo, p53, and APAF1), for a feature vector length of 39. The SimKern simulation uses the same underlying model as the original ODE model with two key differences: 87 of the ODE parameters are marked as uncertain and given Gaussian probability distributions around their true values, and the simulation outputs the time dynamics of the ODEs rather than a classification. 

The {\bf flowering time model} is a set of six ODEs that simulate the gene regulatory network governing the flowering of the Arabidopsis plant \citep{flowering}, and yields a regression problem. 19 mutants are modeled and experimentally validated by the authors. We use those 19 mutational states as well as 34 additional perturbations on the rate parameters to create a varied ground truth sample set. The output of the model is the time to flowering which, following the authors, is set to the time at which the protein AP1 exceeds a particular threshold. For the SimKern model we assume the same model but with uncertainty about the rate parameters. The SimKern simulation output is the time dynamics of the six ODEs. 

The {\bf Boolean cancer model} is a discrete dynamical system of cancer cellular states \citep{booleanNetwork}. Based on the steady state of the system, a sample is labelled as one of three categories: apoptotic, metastasizing, or other. There are no rate parameters since this is a Boolean model. We use the initial state vector (the on/off status of the 32 nodes in the network) as well as mutations of five of the genes (p53, AKT1, AKT2, NICD, and TGF$\beta$) to create a varied sample population with 37 features. In the SimKern simulation, we use a reduced version of the model provided in the original publication. It is unclear how to map the initial conditions from the full model to the initial conditions of the modularly-reduced model, so for all of the modules we randomly choose the mapping, which gives rise to the uncertainty for the SimKern simulations.  The output from the SimKern model (i.e. the data used to form the similarity matrix) is the same classification as from the ground truth model.

The {\bf network flow model} is an optimization problem rather than a simulation. It falls into a subclass of linear optimization models called network flows which are used in a wide range of applications including production scheduling and transportation logistics \citep{lpbook}. The network flow model takes arc costs as inputs, which are the costs of sending a unit of flow through a certain arc in the network. The model then simulates the optimal path of flow along arcs of a directed graph that minimizes the total arc cost along the path. The network is designed in layers and is such that the flow will pass through exactly one of the three arcs in the final layer, which gives us a classification problem (see Figure S2). Changes in arc costs, which represent the features in this model, can lead to changes in the routing of the optimal flow. For the ground truth dataset, we generate samples by varying 12 out of the 80 arc costs. We build two separate SimKern simulations: the better simulation perturbs 23 arc costs, including the 12 costs that were varied to make the ground truth dataset, resulting in a less noisy kernel. The worse simulation varies 21 additional arc costs resulting in a noisier kernel.

\subsection{SimKern simulation -- similarity matrix generation}
\label{simkernGen}
Users must define a model (currently supported languages for the simulation modeling are MATLAB, Octave, and R) which simulates a sample. This simulation procedure, called SIM1 in the python codebase, is used to generate the sample similarity kernel matrix and would be the starting point in an actual application of SimKern. Figure~2 illustrates the SimKern simulation process control.

We assume that there are parameters in this simulation model that we are uncertain about. Let $\theta$ be a vector of these uncertain parameters. We assume we have a random variable description of each of these parameters, which can be very general. For example, a parameter could take the value of 0 or 1 if we have two ways of modeling a particular interaction.
Then, in the simulation, depending on how that random variable gets instantiated, the code uses one of the two parameter values. Alternatively, we might be uncertain about the value of a rate constant, in which case we could use a Gaussian random variable with a specified mean and standard deviation. We assume independence of the random variables $\theta$, but one could also assume a covariance structure.

Each sample $i=1 \ldots N$ is characterized by a feature vector $x_i$, which constitutes sample-specific information that we use to perform the simulations; $x_i$ could be for example a genomic description of patient $i$. For $r=1 \dots R$, where $R$ is the number of trials to run, we instantiate a parameter vector, $\theta_r$. These parameters as well as the sample data $x_i$ are used to run simulation $(i,r)$.

Let $S(x_i, \theta_r)$ (or shorthand, $S_{ir}$) be the simulation output for sample $i$ with uncertainty parameters equal to $\theta_r$. Note that these outputs $S(x_i, \theta_r)$ can be scalars, a classification category, vectors, or any other object. There is no need for these outputs to be the same as what we are trying to predict, $y_i$. We simply assume that given two such outputs, say $S_{ir}$ and $S_{jr}$ for samples $i$ and $j$, we have a way to measure the similarity between them. Let this similarity be given by $z(i,j,r)$. We leave it up to the user to define this function in general (a concrete procedure, for simulations using ordinary differential equations, is given in the Supplementary information).

Finally, the similarity $K(i,j)$ between two samples $i$ and $j$ is the average similarity across the $R$ simulation runs: 
\begin{equation*}
K(i,j) = (1/R)\sum_{r=1}^R z(i,j,r).
\end{equation*}
The above SimKern kernel matrix generation procedure is implemented in Python and is fully described in the Supplementary information.

\begin{figure}
\centering
\includegraphics[trim=0 0 0 0,clip,width=8.35cm]{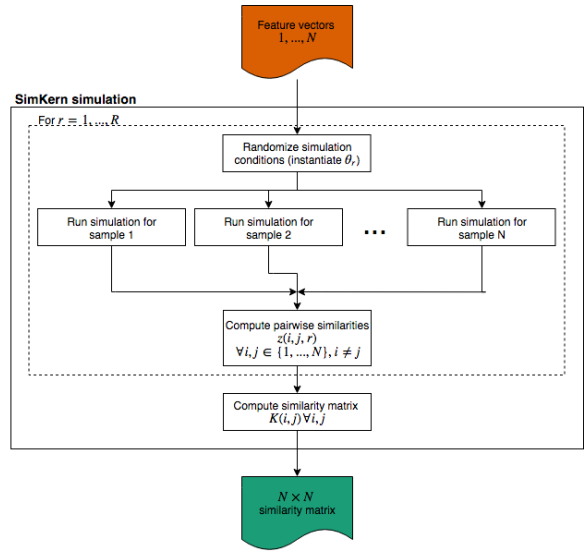}
\caption{Creation of the similarity matrix for use downstream in the machine learning. \label{simkern_sim1}}
\end{figure}

\subsection{Machine learning comparisons procedure}
Figure~1 shows a schematic of the differences in the data processing and machine learning steps for Standard ML and SimKern ML. We compare standard feature-based ML algorithms (orange/top: linear support vector machine (SVM), radial basis function (RBF) SVM, and random forest (RF)) with simulation kernel based methods (green/bottom: kernelized SVM and kernelized RF). We also include results for 1-nearest neighbor (NN) and kernelized 1-nearest neighbor (SimKern NN). As NN-type algorithms are arguably the simplest non-trivial ML algorithms, including these algorithms allows us to understand the distinct contributions of ML algorithm sophistication and simulation-based kernels. 

Since we can generate as many samples as we wish, we train the models and tune the hyperparameters on training and validation datasets which are distinct from the final testing set on which we compute prediction performance metrics (see section \textit{Performance metrics}). The ground truth simulation generates one dataset which is split into three parts (train/validation/test) using the standard proportions 50\%/25\%/25\% \citep[p. 222]{hastie2009elements}.
SVM \citep{ben2010user} and NN algorithms are dependent on feature scaling, therefore features are standardized to the interval $[0,1]$ by subtracting the minimum value and scaling by the range. Categorical features are dummy-coded for SVM and NN algorithms. 
Each ML algorithm is trained on the training data for many hyperparameter configurations and the configuration with the best fit on the validation data is selected. The model given the selected configuration is applied on the test set to compute the performance metrics. See Alg.~1 in the Supplementary information for the details of training, hyperparameter tuning, and testing procedures.

To investigate the performance of simulation-based kernels in scenarios with less data for training, we consider five scenarios in which we train the algorithms on subsamples comprising $s_{1}$, $s_{2}$, ..., $s_{5}$
of the training data. The subsampling percentages are chosen differently per model to highlight the interesting regions of curves that display the performance versus training set size. Table S3 reports the subsampling percentages per model.

\subsection{Performance metrics}
For each of the simulation models, we estimate the generalization performance of an ML algorithm in test data, i.e. data unused for model training, as performance estimates on training data are of little practical value \citep[p. 230]{hastie2009elements}.
The learning tasks per model are either classification or regression. For classification, we consider prediction accuracy, which is defined as
\begin{equation*}
\text{Accuracy} = \frac{\text{true classification count}}{\text{total number of samples}} = \frac{TP+TN}{TP+TN+FP+FN},
\end{equation*}
where $TP$, $TN$, $FP$, $FN$ are the counts of true positives, true negatives, false positives, and false negatives, respectively.
For regression, we consider the coefficient of determination $R^{2}$, which is defined as
\begin{equation*}
R^{2} = 1 - \frac{\text{sum of squared prediction error}}{\text{sum of squares}} = 1 - \frac{\sum_{i}(\hat{y}_{i}-y_{i})^{2}}{\sum_{i}(\bar{y}-y_{i})^{2}}
\end{equation*}
where $y_{i}$ is the outcome for sample $i$, $\hat{y}_{i}$ is the predicted outcome for sample $i$, and $\bar{y}$ is the sample mean of the outcome.
To attain a reliable estimate of the generalization performance, we consider the average test data performance in ten repetitions of a train/validation/test analysis, i.e. repeating training and hyperparameter tuning each time.

\subsection{Standard ML vs. SimKern ML comparison}
For each model, we produce a box plot and/or a line plot that show algorithm performance versus training dataset size for the various ML algorithms in both algorithm groups, Standard ML and SimKern ML. 
\begin{enumerate}
\item Box plots display results for each algorithm separately for the Standard ML (linear SVM, RBF SVM, RF, NN) and SimKern ML algorithms (SimKern SVM, SimKern RF, SimKern NN). The horizontal lines indicate the sample median, the boxes are placed between the first and third quartile ($q_{1}$,$q_{3}$). Outliers are defined as samples outside $[q_{1}-1.5(q_{3}-q_{1}),q_{3}+1.5(q_{3}-q_{1})]$ and are indicated by crosses.
\item Line plots further condense the findings by displaying the median performance metric of the best performing Standard ML and SimKern ML algorithms, excluding NN algorithms in both cases. The best performing algorithm is defined as the algorithm that most frequently produces the highest median performance metric over all five training dataset subsamples. Lines are interpolated for visual guidance.
\end{enumerate}

\subsection{Sensitivity analysis}
To investigate possible factors affecting the SimKern algorithms' prediction performance, we run the following sensitivity analyses:
\begin{itemize}
    \item[] Varying prior knowledge
    \begin{enumerate}
        \item[1)] Radiation model: we examine the results for two kernels which represent different levels of prior knowledge. Both cases utilize the same SimKern simulation, but the higher quality kernel uses the dynamics of only the compartments of the ODE set that are used in the classification of the samples in the initial ground truth simulation. The lower quality kernel uses all ODE equations, therefore not emphasizing the most important ones \citep{pkferranti}.
    \end{enumerate}
    \item[] Varying simulation parameter noise/bias
    \begin{enumerate}
    \item[2)] Network flow model: we generate two kernels for the network flow model. These kernels differ in the number of arc costs that are perturbed and the size of the perturbations (full details in Supplementary information).
    \item[3)] Flowering time model: along with the model that generates the baseline kernel, we study one less noisy, one noisier, and one biased version of the SimKern simulation. The baseline SimKern simulation uses multiplicative Gaussian noise on 34 of the rate parameters, using a mean of 1 and a standard deviation of 0.2. The less noisy model uses stdev=0.1 and the noisier model uses stdev=0.4. For a more radical, and non-centered, departure from the true rate parameters, we also run a model where we multiply each of the same 34 rate parameters with a random variable chosen uniformly from the discrete set $\{0.01,~1,~5,~10\}$.
    \end{enumerate}
    \item[] Varying the number of simulation trials, $R$  
    \begin{enumerate}
    \item[4)] Network flow model: we analyze the effect of additional simulation trials on the prediction performance. We compare the prediction performance of SimKern algorithms when using a similarity kernel based on $R = 3$ simulation trials to the final kernel based on $R = 10$ trials. Furthermore, we track the convergence of the kernel matrix over $R = 10$ trials. 
    \end{enumerate}
\end{itemize}
    

\section{Results}
The general theme that emerges is that, for small training dataset sizes, the methods using the SimKern kernel outperform the Standard ML methods. For larger training sizes, however, the standard methods either approach the SimKern methods or exceed them, depending on the quality of the kernel.

For the radiation model, we see exactly this general pattern (Figure~3). For small training sizes (up to 50 samples), the SVM with the SimKern kernel dominates. We can attribute much of the performance gain to the similarity kernel itself given that the NN algorithm using the same similarity kernel also dominates over the no-prior-knowledge methods for all training sizes shown. The increase in accuracy by the Standard ML algorithms does not yet show signs of saturation by 500 training samples. These box plots are summarized by line plots in Figure 7 (left), which also displays the results of the lower quality SimKern kernel, which was made with the same simulations but without focusing on the most relevant ODEs for the kernel matrix computation.

\begin{figure}
\centering
\includegraphics[trim=40 310 60 280,clip,width=16.70cm]{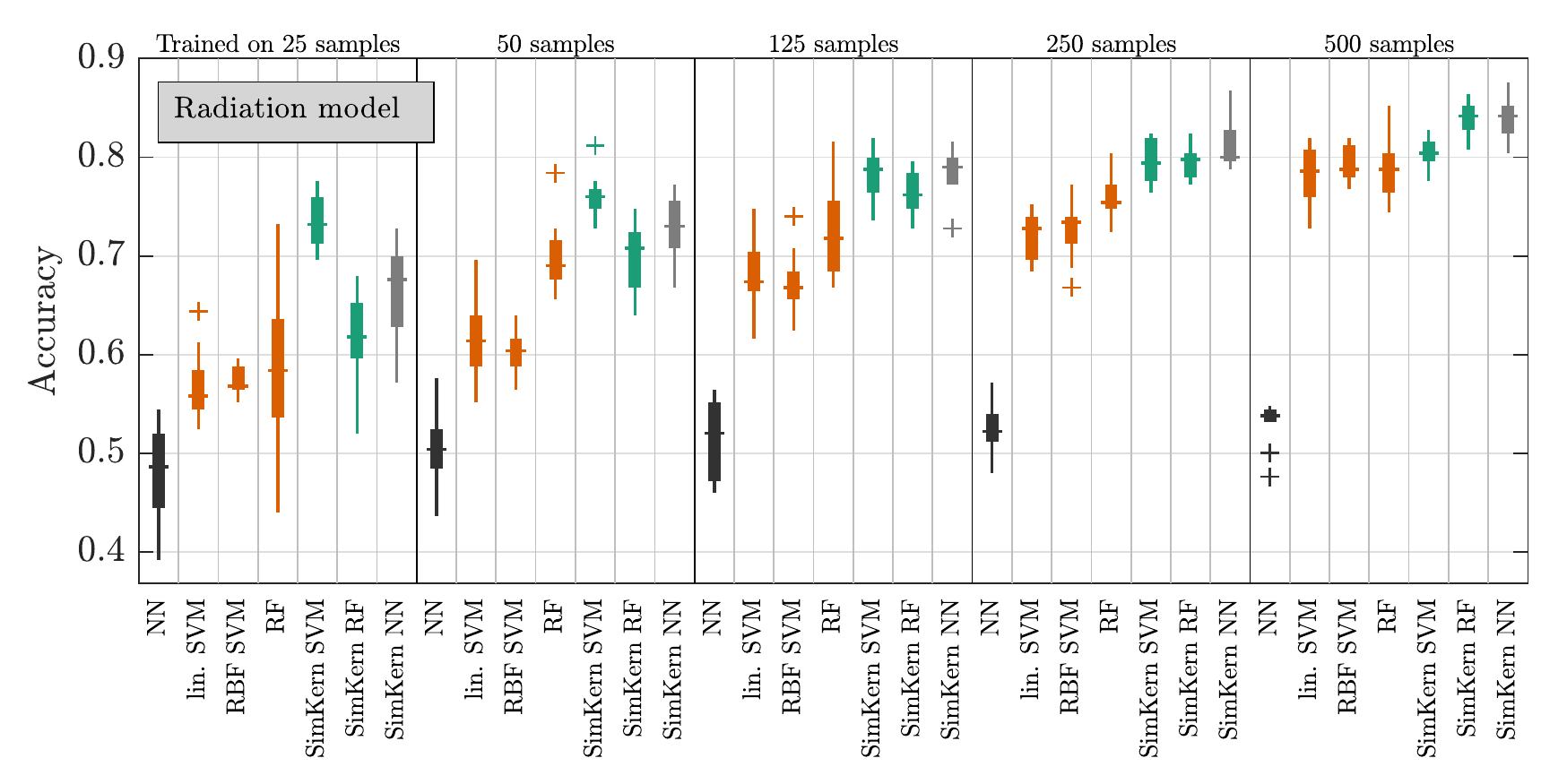}
\caption{Machine learning results for the radiation cancer model. NN = nearest neighbor, RF = random forest, SVM = support vector machine, RBF = radial basis function.}
\end{figure}

\begin{figure}
\centering
\includegraphics[trim=40 310 60 280,clip,width=16.70cm]{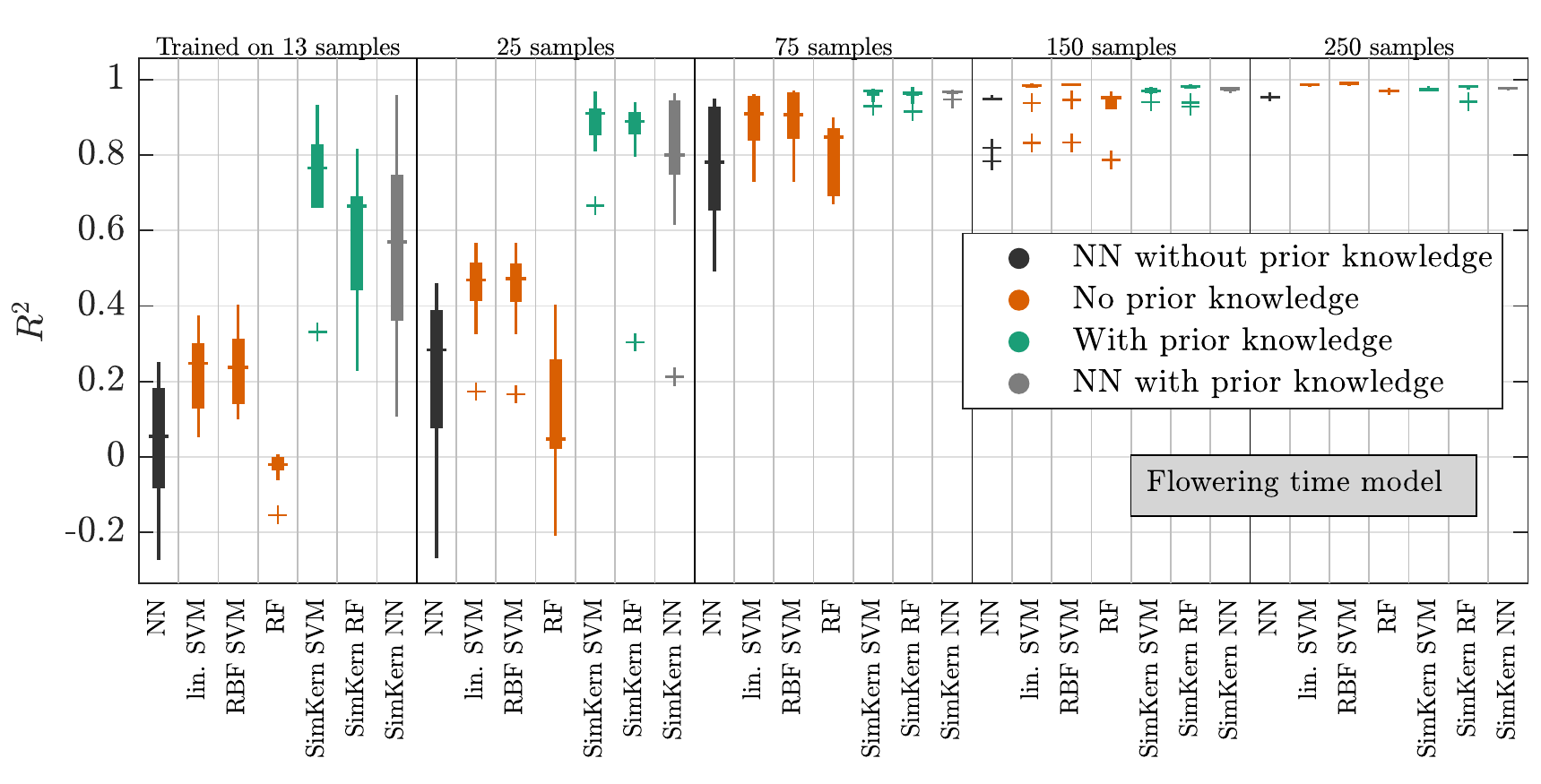}
\caption{Machine learning results for the flowering time model. NN = nearest neighbor, RF = random forest, SVM = support vector machine, RBF = radial basis function.}
\end{figure}

\begin{figure}
\centering
\includegraphics[trim=40 310 60 280,clip,width=16.70cm]{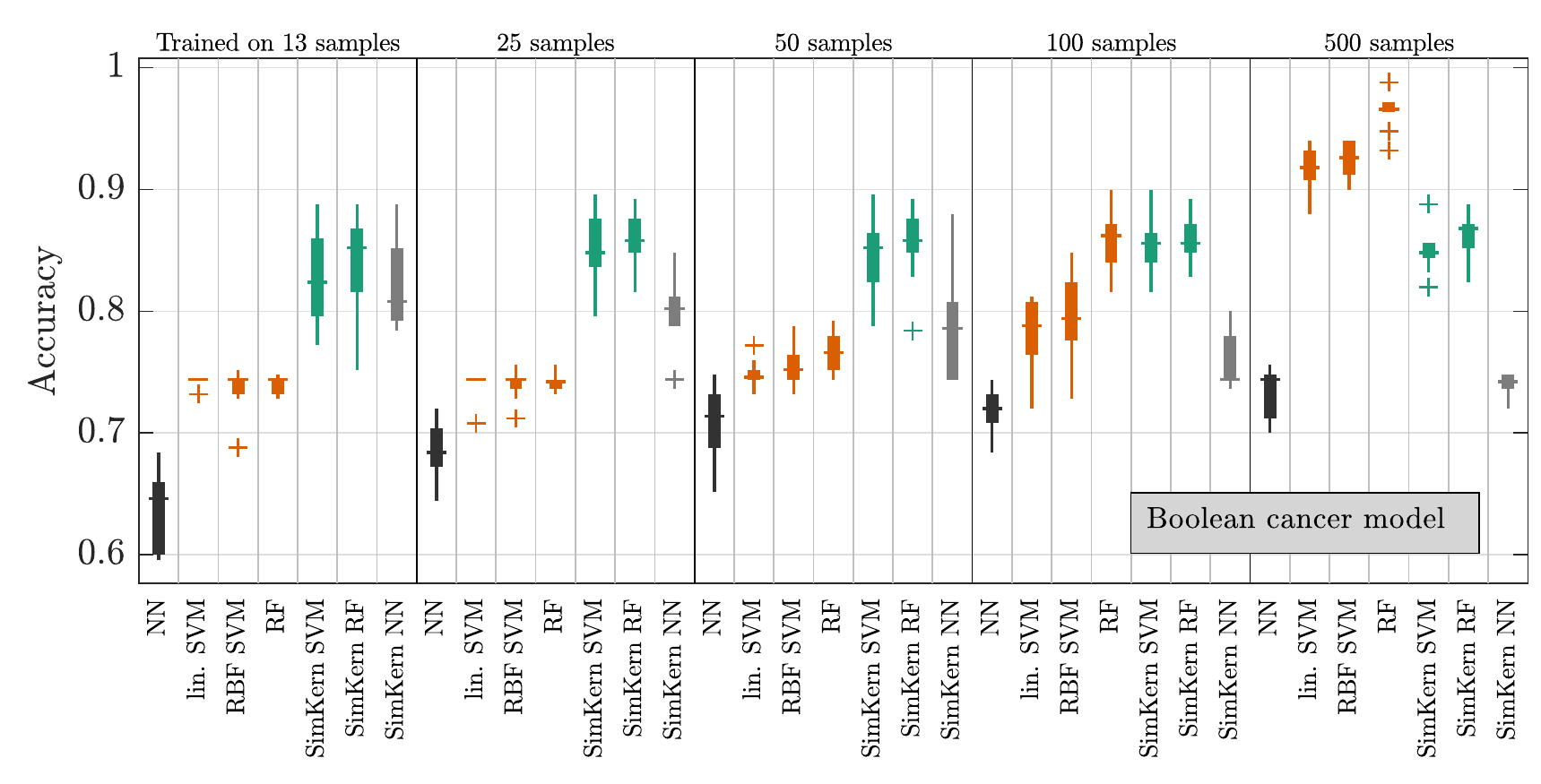}
\caption{Machine learning results for the Boolean cancer model. NN = nearest neighbor, RF = random forest, SVM = support vector machine, RBF = radial basis function.}
\end{figure}

\begin{figure}
\centering
\includegraphics[trim=6 0 0 0,clip,width=16.70cm]{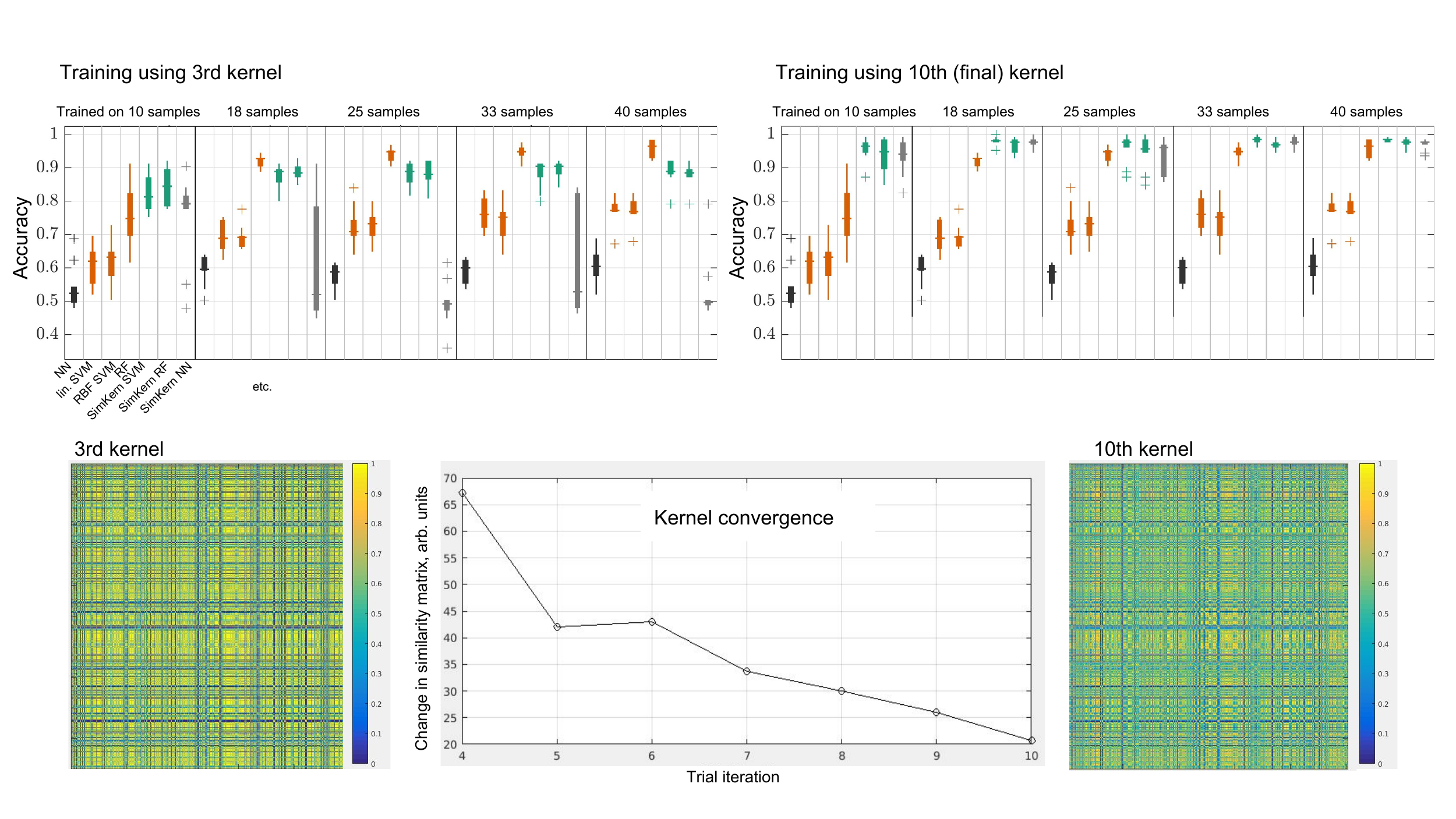}
\caption{Varying simulation trials experiments for the network flow model. The upper two box plots compare the learning accuracy for a kernel from the third of $R=10$ trials versus the final kernel. The kernels themselves are displayed with the same color scale below, and centered at bottom displays the convergence of the kernel (measured using the Frobenius matrix norm) over the ten trials. NN = nearest neighbor, RF = random forest, SVM = support vector machine, RBF = radial basis function.}
\end{figure}

The results of the flowering time model, which also display the clear dominance of SimKern learning for small training data set sizes, show a trend of decreasing variance in predictive performance with increasing training sizes (Figure~4). SimKern learning is strongly dominant up to 75 training samples, after which the two learning styles converge to $R^2 \approx 1$. Another view of the improvement offered by the SimKern method for small training size set sizes is shown by plotting the predicted flowering times versus the actual flowering times, Figure S9.

The sensitivity results obtained by increasing the variance of the (centered) Gaussian noise that was applied to the flowering model's rate parameters display a robustness to these deviations (Figure~8, upper green curves and Gaussian box plots). However, the non-centered noise perturbation analysis shows a clear drop in ML accuracy (Figure~8, dark green dotted line and dark green box plot). With enough training data all SimKern kernels, including the ones with heavy noise, achieve an $R^2$ above 0.95. We call such kernels {\em sufficient}. 

In contrast, the Boolean cancer model kernel is based on a model reduction with additional uncertainty and produces what we call a {\em biased} kernel. There, the SimKern approach produces an accuracy that initially dominates but quickly plateaus to around 85\% and is overtaken by no-prior-knowledge methods when more training data is available (Figure~5). The fact that the kernel learning barely improves with additional data implies that the feature space induced by the simulation kernel is simple enough to be learned by a small amount of samples \citep{effectiveVC}. The kernelized NN method gets worse with more samples, and in general is worse than the other SimKern algorithms, which indicates that the space induced by the biased kernel is less cleanly separable compared to the flowering model case. Above 100 training samples, the no-prior-knowledge RF method is the superior technique.

\begin{figure}
\centering
\includegraphics[trim=100 260 120 280,clip,width=8.35cm]{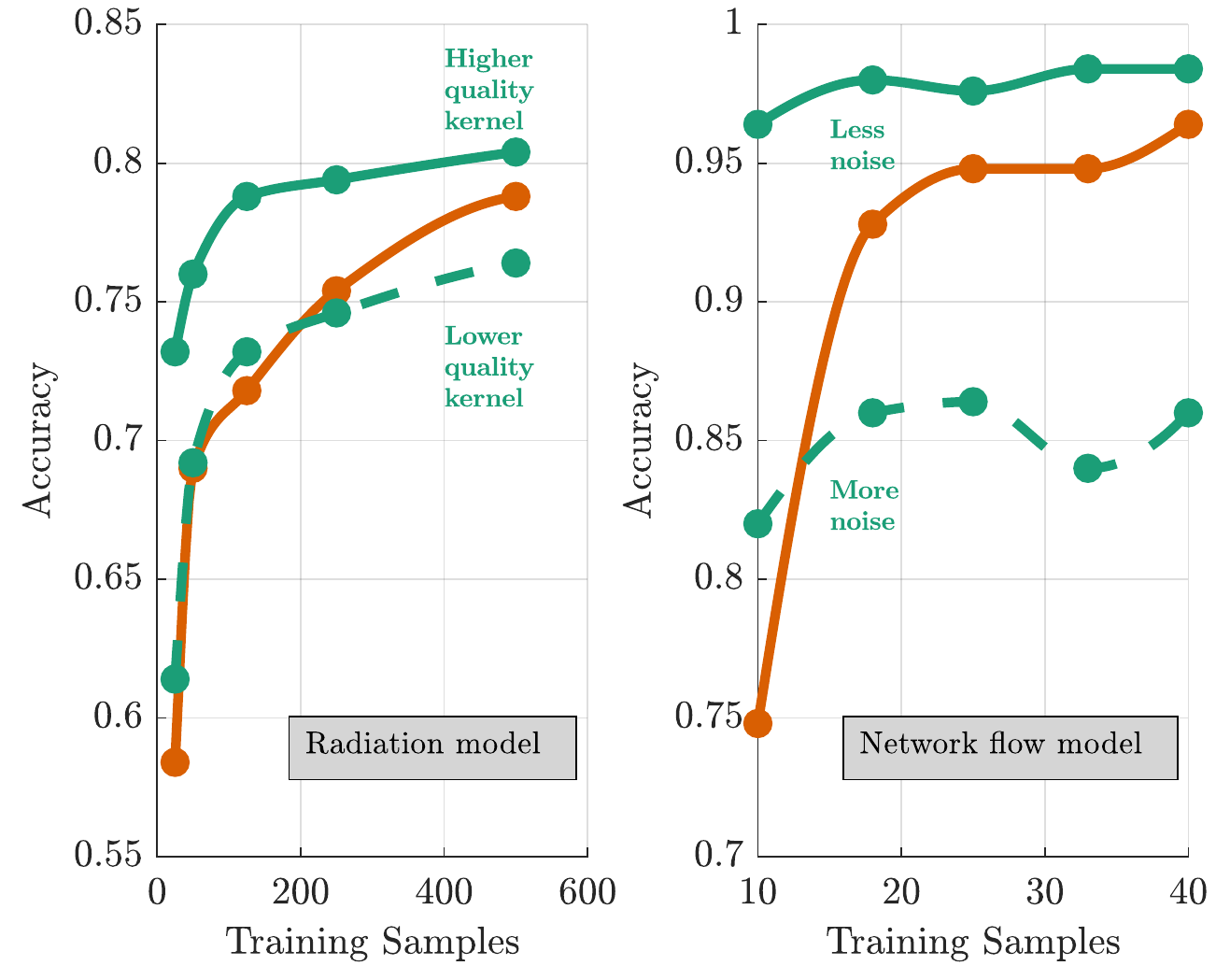}
\caption{Varying prior knowledge experiments for the radiation model (left) and varying parameter noise experiments for the network flow model (right). Performance metrics of SimKern ML based on simulations with less and more prior knowledge (green) and Standard ML (orange). For each line, the best performing algorithm of SimKern ML or Standard ML is selected (see section 2.5 \textit{Standard ML vs. SimKern ML comparison}). Note, the waviness of the less noise case for the network flow model is an artifact of how the data from the box plots was converted into a line plot; the full data, Figure~S7, reveals a flat relationship.}
\end{figure}

\begin{figure}
\centering
\includegraphics[trim=130 80 130 80,clip,width=8.35cm]{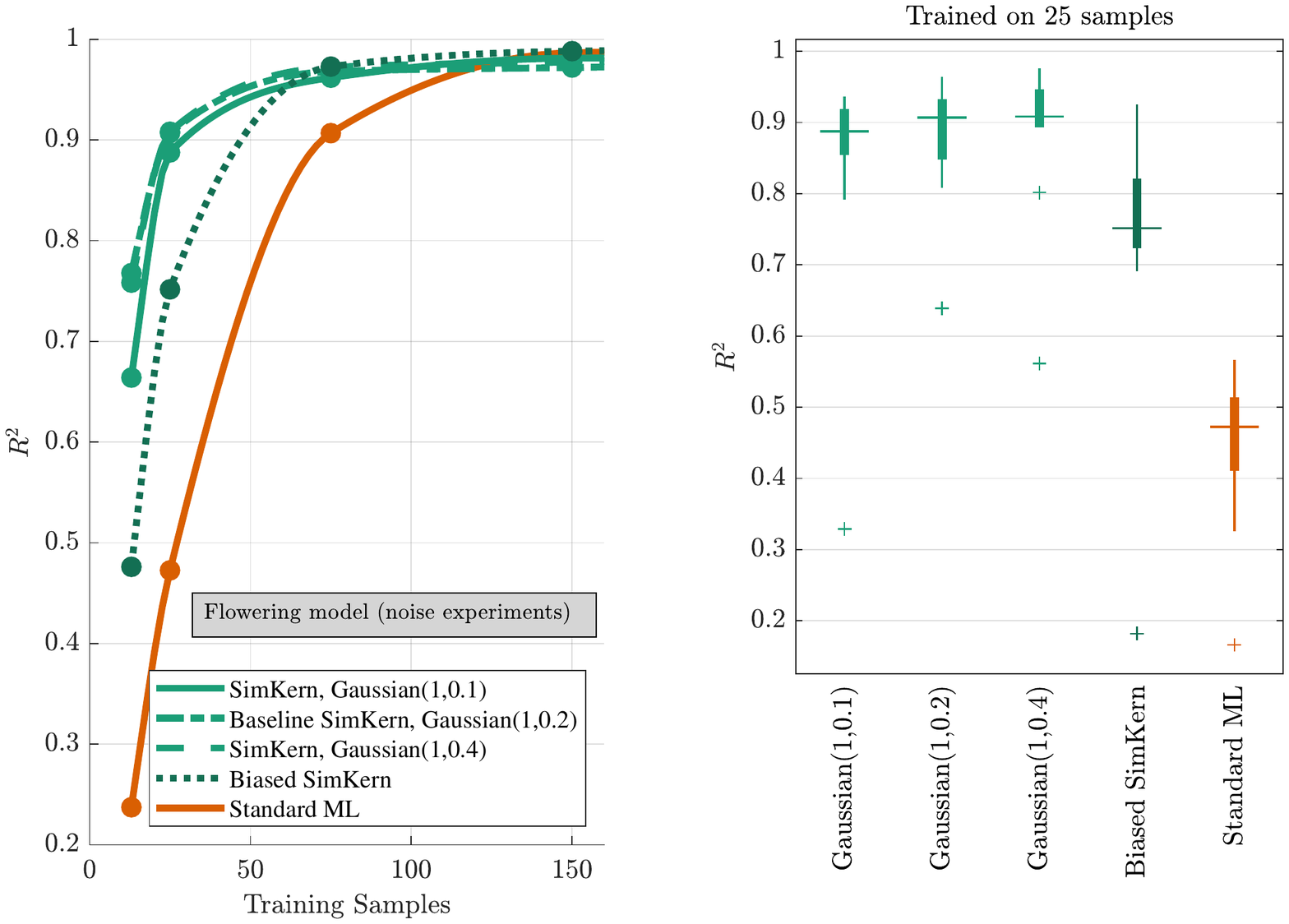}
\caption{Varying simulation parameter noise/bias experiments for the flowering time model. SimKern ML based on simulations with varying parameter noise (green), with parameter bias (dark green), and Standard ML (orange). Left: performance metrics of SimKern ML (green) and Standard ML (orange) trained on up to 150 samples. For each line, the best performing algorithm of SimKern ML or Standard ML is selected (see section 2.5, \textit{Standard ML vs. SimKern ML comparison}). Right: performance metrics box plots for the 25 training sample case.}
\end{figure}

For the network flow problem we evaluate two separate kernels (Figure~7, right) based on different levels of noise in the SimKern simulation: the kernel based on a less noisy SimKern simulation dominates throughout, but even the kernel based on a noisier SimKern simulation is still useful in the very small training set size range. It is doubtful whether one can make general statements about how good a simulation needs to be in order to yield a useful kernel.
However, the intuition that the simulations need only discover the similarity of samples, while not necessarily providing accurate (hence directly useful) simulation results, is described in Figure~S8.

When comparing the individual Standard ML algorithms to the SimKern ML algorithms based on the noisier SimKern simulation (Figure~S7), Standard RF eventually dominates. When comparing algorithms within the Standard ML group, RF is the dominant Standard ML algorithm for the network flow model (Figure~S7) as well as for the Boolean cancer model (Figure~5). For these models, the dominance of RF is likely related to the discrete characteristics of the underlying models. 

The quality of a simulation-generated kernel also depends on the number of trials $R$ that are used to compute the kernel. Figure~6 displays both the convergence of the kernel (bottom) and the improved learning accuracy from the further converged kernel (top), for the less noisy network flow case. We see that the earliest kernel written, kernel three (we chose to not determine similarity kernels below $R=3$), performs noticeably worse than the final kernel. We can also visually observe the differences in the kernels by plotting the 500$\times$500 kernels (Figure~6, bottom left and right). The kernel convergence plot is obtained by taking the Frobenius norms of the difference of the kernel matrices of iteration $i-1$ and $i$, until $i=R(=10)$.

\section{Discussion}
We introduce simulation as a pre-processing step in a machine learning pipeline, in particular as a way to include expert prior knowledge. One can consider simulation as a technique which regularizes data or as a specialized feature extraction method. In either view, the SimKern methodology offers a decomposition of an overall ML task into two steps: similarity computation followed by predictive modeling using the pairwise similarities. This decomposition highlights that to improve the performance of an ML model one can direct efforts into determining better similarity scores between all samples. This is in contrast to the more commonly heard call for ``more data'' to achieve better ML results. Of course, more samples are always desirable, but here we show that, particularly in limited data settings, sizable performance gains can come from high quality similarity scores. 

The decomposition of simulation and machine learning steps also points out their individual contributions. The simulation-based kernel structures the space in which the samples live (or more technically, the dual of the space \citep{kernelML}), and ML finds the patterns in this simplified space. We see that in order to improve machine learning performance we can either improve the kernel or increase the number of samples to better populate the space. For the cases shown here, custom similarity measures show large improvements especially in limited data settings (up to a 20\% increase in classification accuracy and a 2.5 fold increase in $R^2$, depending on the case and the amount of training data used). One could also use the output of the simulations as features for machine learning rather than the additional kernelization step that we employed. Using the simulation outputs directly is related to the field of model output statistics from weather forecasting, where low level data from primary simulations are used as inputs to a multiple regression model which outputs human-friendly weather predictions \citep{MOS}. In our case, we opted for kernelizing the simulation outputs to highlight the fundamental concept of similarity and because a similarity computation is natural when the output of the simulations is a set of time varying entities, e.g., in the case of ODEs. 

Similar in spirit to SimKern, although differing in details, combining simulation and machine learning has been used in physics to predict object behaviour \citep{lerer2016learning,wu2015galileo}. Simulation results are used to train networks to ``learn'' the physics. Varying the simulation conditions during training, called \textit{domain randomization}, is used to improve model generalization \citep{tobin2017domain}. Inversely to the SimKern approach to exploit simulation to enhance ML algorithms, machine learning is also used to correct the inputs to physics simulations \citep{duraisamy2015new}, an idea which is also pursued in the context of traffic prediction \citep{othman2018predictive}.

A novel potential application of the SimKern methodology, one that the authors are currently investigating, involves the prediction of peptides (chains of approximately nine amino acids) binding to a given human leukocyte antigen (HLA) class 1 allele. Current technologies (e.g., \citep{netmhcpan}) predict if a given peptide will bind to a given HLA allele using properties of the amino acids but without using 3D details of the chemical structure of the peptide or information on the structural binding of the peptide and HLA molecule. Computational predictions of binding are considered too difficult at the present time due to the sensitivity of the structural conformations to the detailed chemistry of peptides and the non-covalent interactions \citep{kar2018current}. Nevertheless, simulations could be used to generate similarity scores between peptides, and then the supervised binding data can be used to train a kernelized classification algorithm.

Finally, the use of a SimKern kernel need not be an all-or-nothing decision, since two or more kernels can be combined to yield a single kernel. This allows one to explore the combination of ``standard'' kernelized learning (using uninformed kernels such as linear or RBF) with a SimKern kernel. In the case of a weighted linear sum as the method of kernel combining, one can optimize the weighting vector as part of the training procedure \citep{kernelBook}. Combining kernels allows one to mix traditional feature-based machine learning (which we called Standard ML above) with prior knowledge similarity matrix-based learning.

\section{Conclusions}
It remains to be seen which approaches will be the most fruitful as we make our way towards personalized cancer medicine. Direct testing of chemotherapeutic agents on biopsied patient tissues is a straightforward and promising ``hardware-based'' approach \citep{bh3}. In the machine learning realm, expert feature selection may turn out to be more feasible than the simulation-based kernel methods described in this report. A key question is: can we make simulation-based kernels that--although almost certainly biased--will still be useful (see, e.g., Figure~7)? Progress in detailed biological simulation, such as the full simulation of the cell cycle of the bacterium \emph{Mycoplasma genitalium} \citep{wholecellsim}, the OpenWorm project \citep{openworm}, and integrated cancer signaling pathways for predicting proliferation and cell death \citep{mehdi} offer some encouragement, but cancer influences human biology at all levels, from minute phosphorylations to immune system rewiring. It is thus by no means clear if we are close to simulations that can be useful in this context. However, the magnitude of the problem--both in economic terms and for the number of future patients at stake--suggests pressing forward on all fronts that display conceptual promise.

\bibliography{document}
\bibliographystyle{apalike}
\end{document}



\baselineskip24pt

\maketitle

\tableofcontents

\section{Ground truth and SimKern simulations}

The simulation framework, which handles the generation of ground truth data as well as the SimKern module which performs the simulations and computes the similarity matrix, is written in Python, and supports simulation models written in MATLAB, Octave, and R. It uses text file communication so it could be easily adapted to simulations written in other languages. The Python package, SimKern, is available at github: \url{https://github.com/davidcraft/SimKern}. \textit{We refer to the ground truth simulation as SIM0 and the SimKern simulations as SIM1. This naming convention is also reflected in the Python code base.}

The various code modules are summarized in Table~S1.

\begin{table}[ht]
\begin{center}
\includegraphics[trim=50 55 50 60,clip,width=16cm]{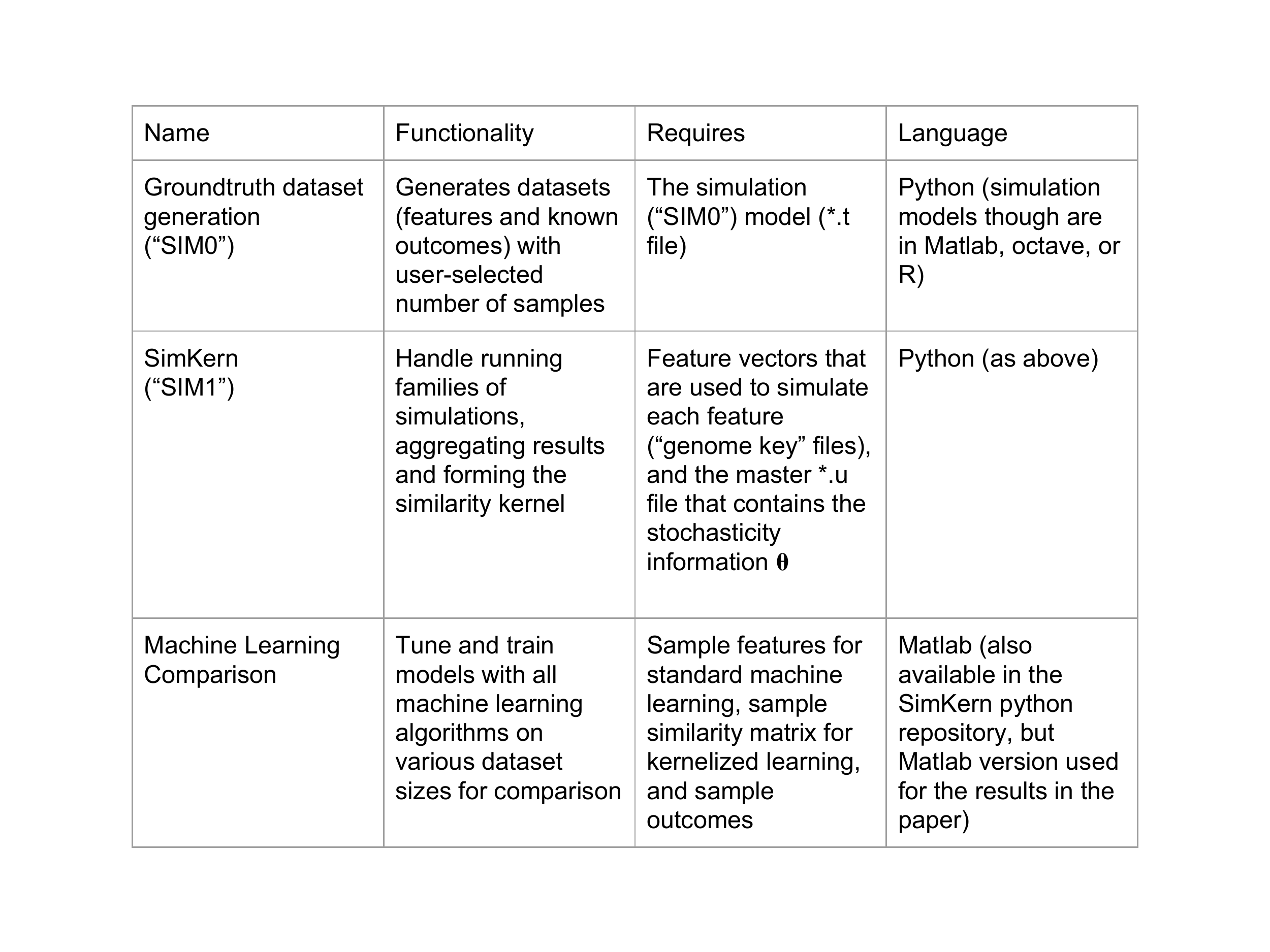}
\end{center}
\label{tab:code_outlines}
\caption{Code module descriptions.}
\end{table}
 
\subsection{Ground truth data generation procedure: SIM0}
A simulation model file used to create a ground truth dataset has the suffix .t. A file used to create the SimKern family of simulations is suffixed with .u (see next section). These model files are in the language of the system used to run the simulations and have entities that are set off by dollar signs. These entities are the parameters to vary from one sample to the next, for the ground truth dataset generation, or from one trial to the next, for the SimKern generation. 

As an example, if different samples may have different values for a rate parameter called $k_1$, a line in the simulation file could read:

\begin{verbatim}
k1 = $gauss(8,2, name=`decayConstant1')$;
\end{verbatim}

The Python code will replace the text set off by the dollar signs with a random variable drawn from a Gaussian distribution with mean 8 and standard deviation 2. In the file storing the sample features that gets written, this feature will be named decayConstant1. This same style is used for both Sim0 and SIM1. The distributions that are allowed, and more usage details, are given in the manual on the SimKern github repository.

If the simulation package to use is MATLAB, the Python package allows a direct process hook via a MATLAB-Python API provided by MathWorks. This speeds up the overall runtime by not requiring the expensive startup time of MATLAB for every run.

\begin{figure}[!ht]
\centering
\includegraphics[trim=0 0 0 0,clip,width=4.17cm]{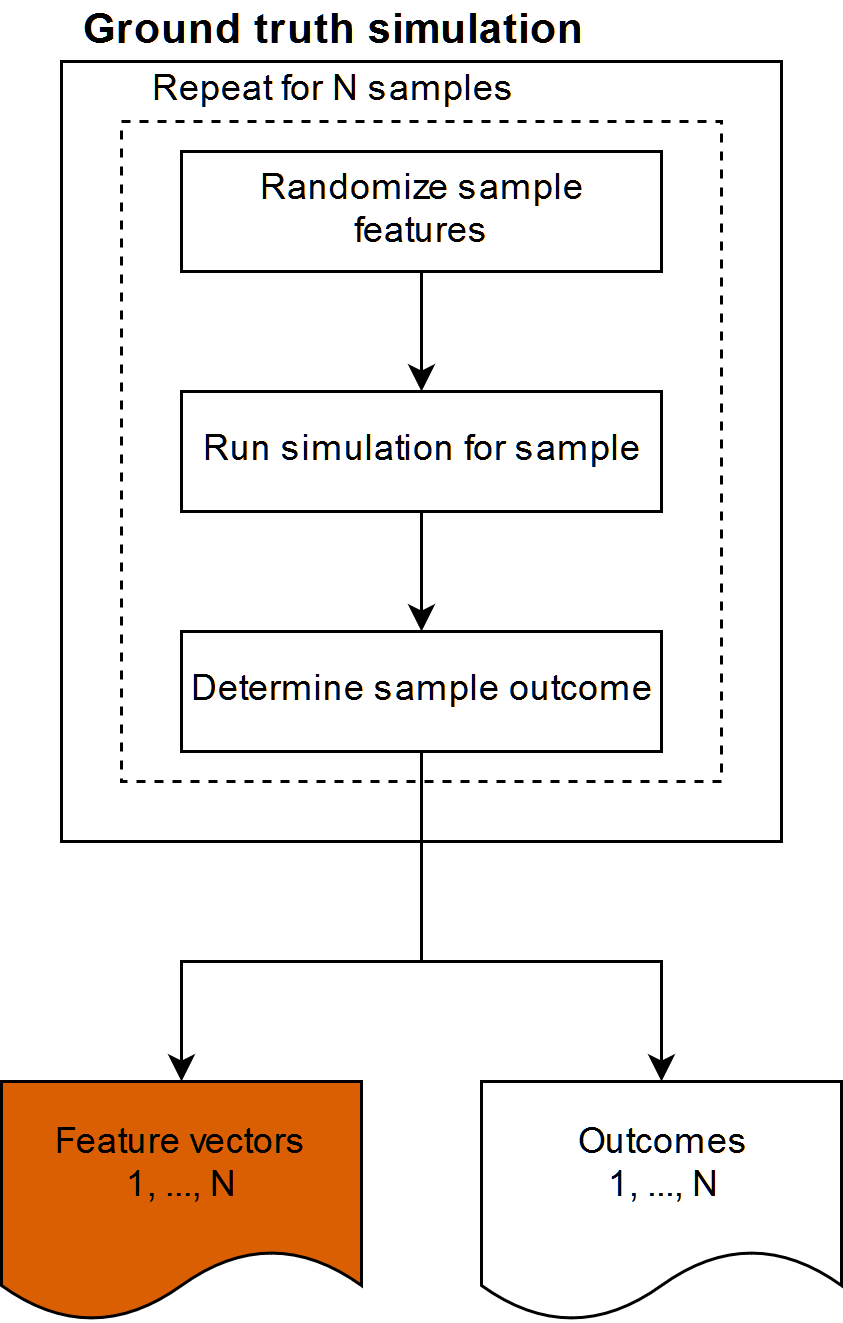}
\caption{Ground truth data generation procedure, SIM0.} \label{ground_truth_sim}
\end{figure}

Let $N$ be the number of samples we generate for the SIM0 dataset. Let the feature vectors (the parameters that make the samples different from each other) be given by the vectors $x_i$, $i=1 \ldots N$. Each $x_i$ vector is a vector of length $p$, where we are following the standard machine learning notation where $p$ equals the number of features.  Let $y_i$ denote the outcome of the simulation, which could be a category (e.g. alive or dead) or a real number. Since we generate these outputs via a simulation, viewing that simulation as a function $S^0$ we can write $y_i = S^0(x_i)$. The ground truth data generation procedure is depicted in Figure~\ref{ground_truth_sim}.

The $x$ data get written to Sim0Genomes.csv and the $y$ data to Sim0Output.csv (the term genome is used since the use case that provides the motivation for this software is machine learning for biological systems where the feature vector is based on genomics). Separate files, called genome keys, are written out for each sample for use in the SIM1 runs.


\subsection{Similarity kernel generation: SIM1}

The main document describes the similarity matrix computation. 
The python software handles writing out and running the individual $(i,r)$ run files, using the .u file as the template. This .u file must reference another file which specifies the parameters from the SIM0 run that make each individual sample $i$ distinct. This file is called genome1\_key (the ``1'' is replaced automatically by the SIM1 python code with the sample number $i$).

The output of this procedure is the similarity matrix, given in a file called SimilarityMatrixfinal.csv. A similarity matrix is also written after every trial (from the third trial onward; similarity matrices before the third trial are considered not converged yet and so are not written out).

\subsubsection{Similarity as measured by closeness of ODE solutions}

A typical setting for a SIM1 run will be the simulation of a set of ordinary differential equations (ODEs). In this context, the similarity between population members $i$ and $j$, for simulation $r$, can be a measure of how close the overall time dynamics for $i$ are to the time dynamics of $j$, e.g., represented by the mean squared error over discrete time points. More specifically, assume the ODE simulation contains $E$ different entities (e.g. protein levels), in other words $E$ ODEs. Let us further assume that the simulation program outputs the levels of these entities at a given set of times, $t_1,~t_2, \ldots,t_k,\ldots t_T$. Let $L^i_r(e,k)$ be the level of ODE entity $e$ at time $t_k$, for population member $i$ under simulation $r$. Since the ODE equations may be of different magnitudes, we will normalize each pair being compared by the maximum level that either ever takes over the time course (we are implicitly assuming the ODEs solutions are always non-negative, this would have to be modified for negative levels). For the pair of samples $(i,j)$ and for entity $e$ in simulation run $r$,
the maximum value $M$ is given by:
$$
M(i,j,e,r) = \mathrm{max} [ \mathrm{max}_{k} L^i_r(e,k),~\mathrm{max}_{k} L^j_r(e,k) ] 
$$

With these definitions, we can write

\begin{equation}
z(i,j,r) = 1 - \frac{1}{E\cdot T}\sum_{e=1}^E \sum_{k=1}^T  \left( \frac{L^i_r(e,k) - L^j_r(e,k)}{M(i,j,e,r)} \right) ^2 
\end{equation}

Finally, in addition to normalizing the ODE solutions to a maximum value of 1, the user may want to
weight the different entities $e$ to express the prior knowledge that some entities are more important
for similarity considerations than others. Let $ 0 \le w_e \le 1$ be user-defined weights and then we
have:
\begin{equation}
z(i,j,r) = 1 - \frac{1}{E\cdot T}\sum_{e=1}^E w_e \sum_{k=1}^T  \left( \frac{L^i_r(e,k) - L^j_r(e,k)}{M(i,j,e,r)} \right) ^2 
\end{equation}

\section{Machine learning details}

\subsection{Machine learning algorithm comparisons procedure}

Machine learning (ML) was conducted in MATLAB (MathWorks, Natick, MA, USA) using the libSVM package for all SVM models \cite{chang2011libsvm}. The python SimKern codebase also provides routines for the machine learning runs. Alg 1. outlines the experimental design to tune the hyperparameters and then estimate performance metrics for each ML algorithm. Although the algorithm initially splits a dataset into three pieces--50\% for training, 25\% for validation, and 25\% for final accuracy assessment--the training subset is further subsampled to assess how accuracy depends on the amount of training data for the various models and machine learning algorithms. The same experiment is repeated for each dataset. The procedure is outlined below and explained in detail in the subsequent subsections.

\begin{center}
\colorbox[gray]{0.95}{
\SetAlgoLined
\SetNlSty{textbf}{}{:}
\begin{algorithm}[H]
\SetAlgoLined

load data of the ground truth data simulation\;
load similarity matrix of the SimKern simulation\;
shift and rescale features to $[0,1]$\;
dummy-code categorical features for SVM algorithms\;

\For{{\upshape repetition} $i=1:10$}
{
randomly sample 50\% of all rows as training data (stratify samples if it is a classification problem)\;

randomly sample 25\% of all remaining rows as validation data (stratify samples if it is a classification problem)\;
assign the remaining rows as test data\;
\ForEach{{\upshape subsampling percentage} $s\in\{s_{1},s_{2},...,s_{S}\}$ }
{
randomly subsample $s$ of all training rows as training data (stratify samples if it is a classification problem)\;
\ForEach{{\upshape algorithm} $a \in A$}
{
\ForEach{{\upshape hyperparameter configuration} $h_{a} \in H_{a}$} 
{
train algorithm $a$ with hyperparameter configuration $h_{a}$ on training data\;
predict outcomes for validation data\;
compute performance metric on validation data predictions\;
}
select hyperparameter configuration $h_{a}^{\ast}$ with best validation performance metric\;
select algorithm $a$ trained with hyperparameter configuration $h_{a}^{\ast}$\;
predict outcomes for test data\;
compute performance metric on test data predictions\;
}
}
}
\label{alg:expdes}
\end{algorithm}
}
\end{center}
\noindent {\bf Alg. 1.} Experimental design to estimate ML performance (this algorithm is executed independently on each dataset). $A$ is the set of ML algorithms used. $s_i$ subsampling percentages vary by model in order to home in on the most relevant part of the curve which represents accuracy versus amount of training data, see Table~S3.

\subsubsection{Stratification}

 For the classification models, the data is split while approximately stratifying for classes. Stratification of classes in training, validation, and test data ensures stability in the estimation process. Consider the case where random sampling led to an unusual distribution of classes in training and validation data. Consequently, the test data would very likely have a class distribution different than the training data. Classifiers not correcting for class imbalance (default RF and default SVMs) that are trained on this training data would perform worse on the test data. Since we want to estimate generalization performance, i.e. performance on the general population with a class distribution estimated by the class distribution in the full dataset, we stratify classes in training and test data.

\subsubsection{Hyperparameter tuning}

The performance of the studied ML algorithms is dependent on algorithm-specific hyperparameters (HP) whose optimal values for generalization performance are not known {\em a priori}. HPs are tuned by a grid search: for a selection of values per HP, the algorithm is trained on the training data and evaluated on the validation data for each possible HP combination. The HP combination with the best performance metric in the validation data is selected. Table S2 lists the HPs that are tuned, their ranges, and values on the search grid for each algorithm. Values are partly determined from existing literature or chosen experimentally. HPs not mentioned here are set to default values. Values for SVM parameters are partially taken from \cite{ben2010user}. For RF, the number of trees is fixed at 100. 
While \cite{breiman2001random} did not limit the number of terminal nodes in a tree, \cite{duroux2016impact} provide empirical evidence in favor of tuning. Therefore, we tune the maximal number of splits allowed in a tree. Tuning grid boundaries have been extended manually to reduce the number of cases where the tuning procedure selects HP values on the grid boundaries, which would suggests that better HP values might be found outside the grid.

\begin{table}[htbp]
\begin{center}
\begin{tabular}{cccc}
    \toprule
     \textbf{Algorithm} & \textbf{HP} & \textbf{Range} & \textbf{Values on grid} \\
     \midrule
     \multirow{2}{*}{\shortstack{linear SVM\\ \& SimKern SVM}} & $C$ & $[0,\infty]$ & $\{10^{-12},10^{-11},...,10^{12}\}$ \\
     & $\varepsilon$ & $[0,1]$ &  $\{10^{-5},10^{-4},...,10^{-1},0.25,0.5,0.75,1\}$ \\
         \hline
     \multirow{3}{*}{RBF SVM} & $C$ & $[0,\infty]$ & $\{10^{-12},10^{-11},...,10^{12}\}$ \\
             & $\gamma$ & $(0,\infty]$ & $\{10^{-15},10^{-14},...,10^{1}\}$ \\
             & $\varepsilon$ & $[0,1]$ &  $\{10^{-5},10^{-4},...,10^{-1},0.25,0.5,0.75,1\}$ \\
         \hline
     \multirow{2}{*}{\shortstack{RF\\ \& SimKern RF}} & \emph{n. feat.} & $[1,\infty]$ & $\{1,\lfloor (1+\sqrt{p})/2\rfloor,\lfloor\sqrt{p}\rfloor,\lfloor (\sqrt{p}+p)/2\rfloor,p\}$ \\ 
      & \emph{n. splits}  & $[1,(n-1)]$ & $\lfloor\{0.05,0.1,0.2,0.3,0.4,0.5,0.75,1\}n\rfloor$ \\  
     \bottomrule

\end{tabular}
\end{center}
\label{tab:hp}
\caption{Hyperparameter tuning per algorithm. $C$ is the weight corresponding to training set error in the SVM objective. $\varepsilon$ (only used for SVM regression) determines the width of the margin enclosing the separating hyperplane in SVM regression. $\gamma$ is a parameter of the RBF kernel $K(x,y) = \exp{\left(-\gamma||x-y||^{2}\right)}$. \emph{n. feat.} is the number of randomly sampled features compared at each split in a tree. \emph{n. splits} is the maximal number of splits per tree, grid values exceeding the interval $[1,(n-1)]$ are truncated to the boundary. $n$ is the number of training samples, $p$ is the number of features.}
\end{table}

\begin{table}[htbp]
\begin{center}
\begin{tabular}{cccccc}
    \toprule
     \textbf{Model} & {\boldmath$s_{1}$} & {\boldmath$s_{2}$} & {\boldmath$s_{3}$} & {\boldmath$s_{4}$} & {\boldmath$s_{5}$}  \\
     \midrule
    Radiation & 5\% & 10\% & 25\% & 50\% & 100\% \\
    Flowering & 5\% & 10\% & 30\% & 60\% & 100\% \\
    Boolean & 2.5\% & 5\% & 10\% & 20\% & 100\% \\
    Network & 4\% & 7\% & 10\% & 13\% & 16\% \\
     \bottomrule
\end{tabular}
\end{center}
\label{tab:subsamp}
\caption{Subsampling training percentages per model.}
\end{table}

\subsection{Machine learning algorithms}
\begin{figure*}[!ht]
\centering
\includegraphics[trim=0 0 0 0,clip,width=16.70cm]{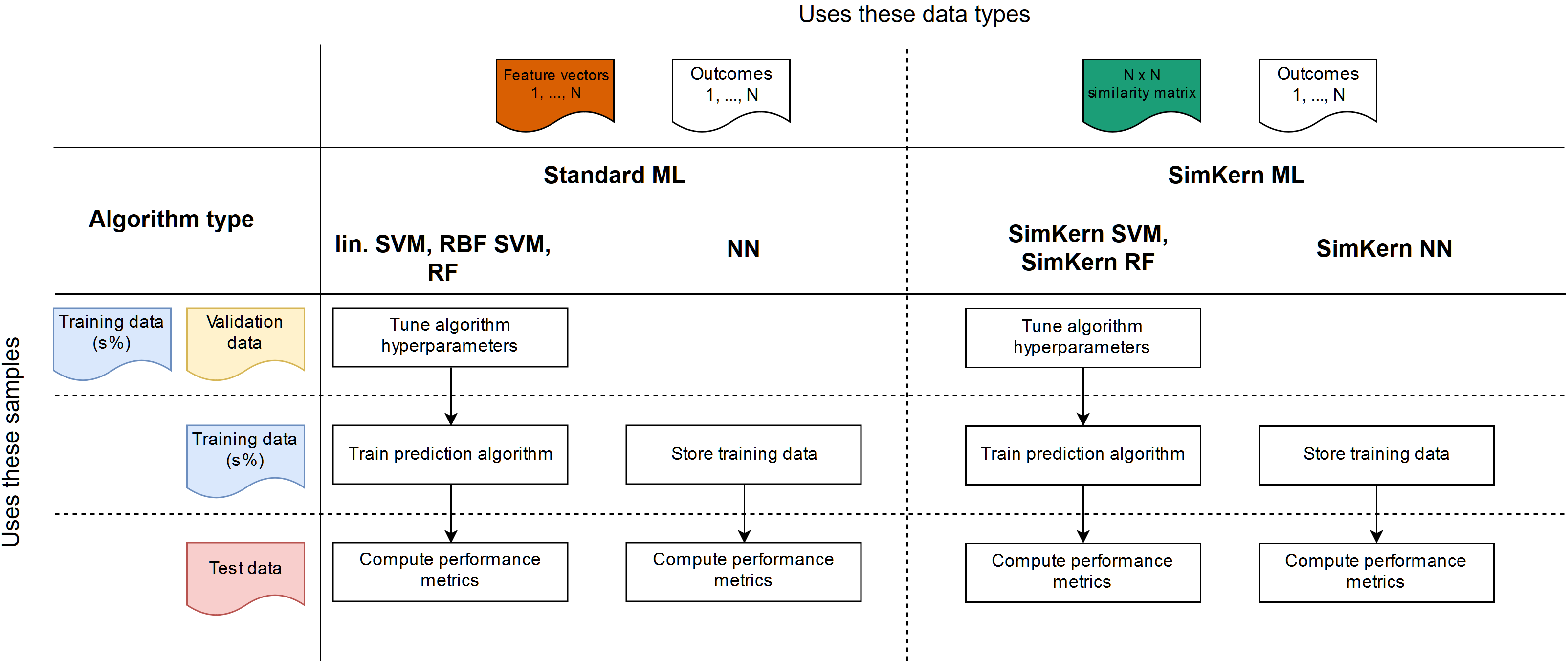}
\caption{An overview of the data handling procedures for the various machine learning algorithms used. SVM=support vector machine, RBF=radial basis function, ML=machine learning, NN=nearest neighbors, RF=random forest. \label{learning}}
\end{figure*}

We compare standard machine learning algorithms that use the ground truth feature vectors (Standard ML algorithms) to ML algorithms that use the SimKern kernel matrix (\emph{SimKern} ML algorithms), see Figure~\ref{learning}. For the Standard ML learning, we utilize three established ML algorithms: linear SVM \cite{cortes1995support}, radial basis function (RBF) SVM, and random forest (RF) \cite{breiman2001random}. For SimKern learning, we use SVM with the similarity matrix as a custom kernel 
(note that for the SVM algorithm the kernel matrix, also known as the Gram matrix, has to be symmetric positive definite, which in all of our models is the case, and indeed is required by the libSVM software) and the random forest algorithm with the similarity matrix as the feature matrix input \cite{sathe2017similarity}. This random forest, called SimKern RF, classifies new samples according to their similarities with training samples.
Additionally, we compute nearest neighbor predictions to compare to the more advanced machine learning algorithms. For the Standard ML case, we use a 1-NN algorithm on the SIM0 feature vector. For the SimKern case, we use the label of the most similar distinct training sample according to the similarity matrix. We label this approach \emph{SimKern NN}.

\section{Model descriptions}

Table~S4 gives a summary of the machine learning problem sizes, number of features, and other attributes, for the four models.

\begin{table}[htbp]
\begin{center}
\begin{tabular}{cccccccc}
    \toprule
     \textbf{Model} & \textbf{Class 1} & \textbf{Class 2} & \textbf{Class 3} & \textbf{Class 4} & $n$ & $p$ & $R$\\
     \midrule
    Radiation & 27.5\%  & 23\%  & 44.2\% & 5.3\% & 1000 & 39 & 20\\
    Flowering & - & - & -  & - & 500 & 35$^*$ & 5 \\
    Boolean & 62.5\% & 9.3\% & 28.2\%  & - & 1000 & 37 & 20\\
    Network & 61.6\% & 18.2\% & 20.2\% & - & 500 & 12 & 10\\
     \bottomrule
\end{tabular}
\end{center}
\label{tab:classdistribution}
\caption{Numerical information for the four models. Class distribution per model for the ground truth (SIM0) dataset. Note that the Flowering model has continuous outcomes (i.e. flowering time) and the Boolean and Network models have only three classes. Classes (in order 1, 2, 3, 4) for the Radiation model are apoptosis, repaired and cycling, mitotic catastrophe, and quiescence. For the Boolean cancer model they are apoptosis, metastasis, and other. For the Network model they are simply which of the exit arcs the optimal solution flows through. $n$ is the number of samples generated for the SIM0 ground truth dataset, $p$ is the number of features in the ground truth dataset, and $R$ is the number of trials run in the SimKern step. $^*$For the flowering model one of the features is a categorical variable of 19 classes, representing 19 different mutational states. Thus if one-hot encoded this would lead to an additional 19 features.}
\end{table}

\subsection{Radiation model}

The radiation model is built up as four connected modules. We opt to not simulate the cell cycle and instead focus on the chain of events that happens after radiation damages a cell's DNA: DNA repair (modeled at a high level), p53-based transcription factor control, cell cycle arrest, and apoptosis, see Figure~S3. Although highly simplified, this model recapitulates the idea that the inter-connected dynamics of these processes determine cell fate after radiation damage.

Tuning this model to reflect the behavior of an actual cell line is very large task, and probably not possible in any realistic way, since the genes (proteins) chosen to be in the model are but a small subset of the proteins involved in a DNA repair and cell cycle control cascade. However, even without validated rate constants chosen, the model provides a numerical instance of a complex system, based on known biology, where different modules (biochemical processes) are involved in determining the fate of a cell subject to an external stimulus. We hand tuned the parameters of the base model. There are many parameters to choose from, and our choices were from manual explorations which led to a set of parameters that led to diverse system behavior (some samples ending in apoptosis, others in cell cycle arrest, etc.).

\begin{figure}[!ht]
\centering
\includegraphics[trim=0 0 0 0,clip,width=14cm]{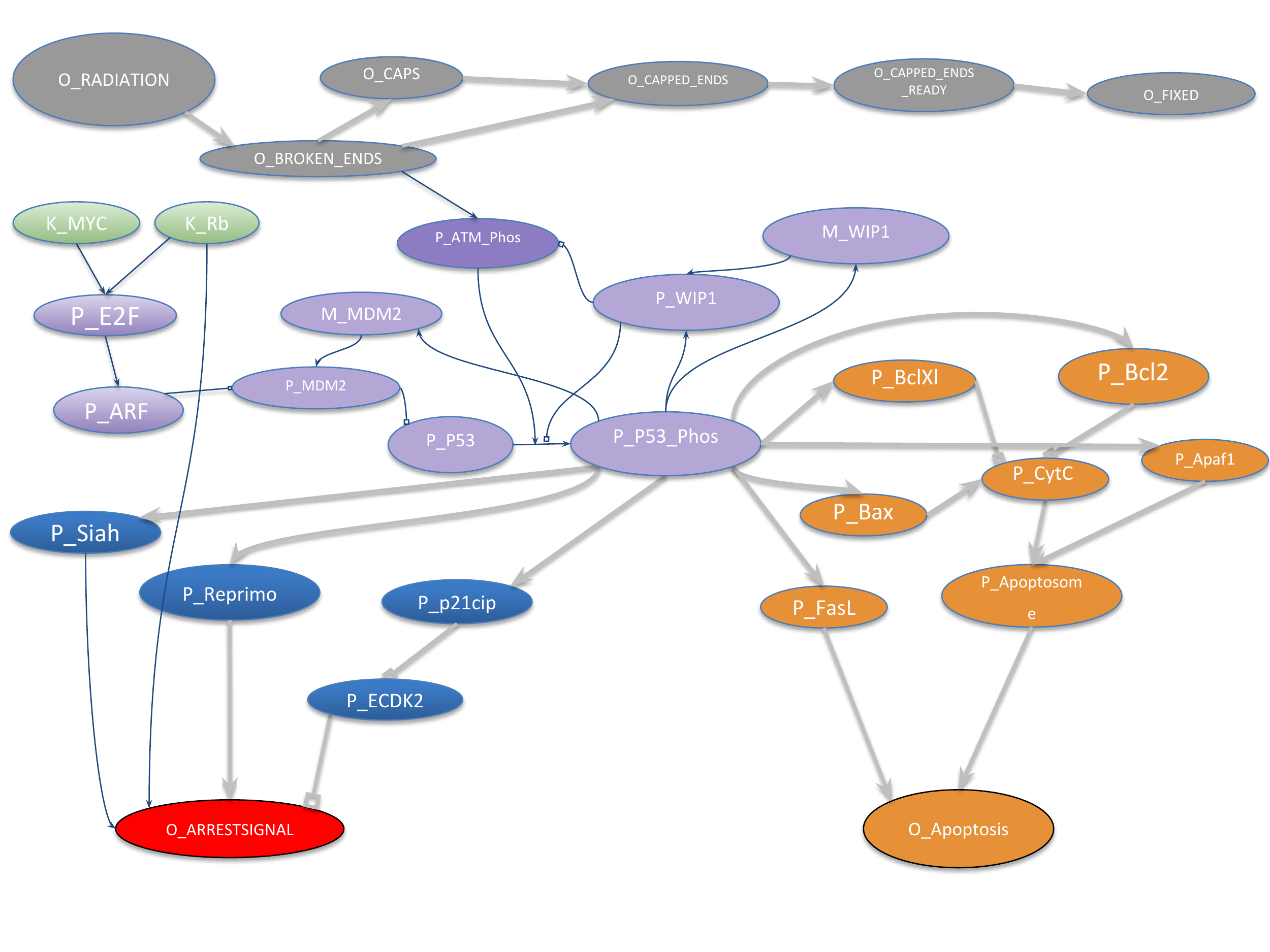}
\label{radcancer}
\caption{ A model of entities and processes involved in cell fate decision following radiation. The gray nodes depict
  the process of DNA breakage and repair. DNA breaks send signals via ATM to the p53-MDM2-ARF module (purple),
  which in turn sends both apoptotic signals (orange) and cell cycle arrest signals (blue). The cancer genes MYC and Rb (green) are modeled as fixed parameters rather than time varying entities. The first letters of each oval have the following meanings: P = protein, M =  mRNA, K = rate constant, O = other. Phos stands for phosphorylated.}
\end{figure}

The p53-MDM2 transcription regulatory control circuit comes from \cite{elias}.
We use the single compartment version of the model, where
the specific location of molecules (nucleus versus cytoplasm) is ignored. Radiation damage
affects this circuit via the ATM kinase pathway, which increases the phosphorylation and hence stability of p53. p53 then goes on to be a transcription factor for apoptosis and cell cycle arrest genes.

Cell specific alterations (mutations, amplifications, deletions) for MYC, RB1, and p53 interact to influence how the p53-MDM2 circuit behaves, which in turn affects the behavior of the downstream processes of cell cycle arrest and apoptosis. The number of cell cycle controls in an eukaryotic cell is large. Rather than attempting to model most of them, we choose a few overlapping controls to create a model that creates a challenging machine learning problem.

Apoptosis is modeled as the competition between pro-apoptotic (BAX, FasL) and anti-apoptotic proteins (BCL-2, BCL-xl). Apoptosis occurs if the apoptosome is formed (a combination of cytochrome c and APAF-1, which together release caspases from the mitochondrial membrane) or via the extrinsic Fas/FasL pathway.

The detailed mathematical model is given next.  In the ODE equations as written below, we use a generic ``$k$'' for ODE constants, to reduce clutter. For the full details, we refer the reader to the MATLAB code. 

Phosphorylated nuclear p53 protein tetramerizes to form its active transcription factor state. For convenience we define the p53 tetramerized term as:
\begin{equation}
p53tt = (MUT_{p53}*pP53NucPhos^4)
\end{equation}
The mutation coefficient $MUT_{p53}$ is a uniform random variable between 0 and 1, reflecting the idea that there are a large number of p53 mutations that potentially affect the tetramerization in varying ways. 

The full ODE model is given here:
\begin{align*}
\dot{o Radiation} &= -k * oRadiation\\
\dot{o Broken Ends} &= k * oRadiation - k * oBroken Ends * oCaps\\
\dot{o Caps} &= \min{((k * o Broken Ends),k_5)} - k * o Broken Ends * o Caps - k * oCaps\\
\dot{o Capped Ends} &= k * o Broken Ends * o Caps - k * o Capped Ends\\
\dot{o Capped Ends Ready} &= k* o Capped Ends - k * o Capped Ends Ready\\
\dot{o Fixed} &= k* oCapped Ends Ready\\ 
\dot{p P53Nuc} &= k + k * p WIP1Nuc * \dfrac{pP53NucPhos}{ + pP53NucPhos} - \\
~ & k*pMDM2Nuc * \dfrac{pP53Nuc}{k*pP53Nuc}-k*pATMNucPhos\dfrac{pP53Nuc}{k*pP53Nuc} - \\ & k*pP53Nuc\\
\dot{p MDM2Nuc} &= k* mMDM2Nuc -pMDM2NUC - \\
 & MUT_{arf} * k*pARF*pMDM2Nuc\\
\dot{m MDM2Nuc} &= k + k*\dfrac{p53tt}{k^4 + p53tt}-k * mMDM2Nuc-k*mMDM2Nuc\\
\dot{p P53NucPhos} &= k* pATMNucPhos * \dfrac{pP53Nuc}{k*pP53Nuc} - k * pWIP1Nuc * \dfrac{k*pP53NucPhos}{k+pP53NucPhos}\\
\dot{p WIP1Nuc} &= k * mWIP1Nuc -k pWIP1Nuc\\
\dot{m WIP1Nuc} &= k +k*\dfrac{p53tt}{k^4+p53tt}-k*mWIP1Nuc-k*mWIP1Nuc\\
\dot{p ATMNucPhos} &= 2*k*oBroken Ends*\dfrac{\dfrac{k-pATMNucPhos}{2}}{k+\dfrac{k-pATMNucPhos}{2}} - \\
 & 2*k*pWIP1Nuc*\dfrac{ATMNucPhos^2}{k+pATMNucPhos^2}\\
\dot{p Bcl2} &= k*\dfrac{p53tt}{k+p53tt}-k*pBcl2\\
\dot{p BclXl} &= k*\dfrac{p53tt}{k+p53tt}-k*pBclXl\\
\dot{p FasL} &= k*\dfrac{p53tt}{k+p53tt}-k*pFasL\\
\end{align*}
\newpage
\begin{align*}
\dot{p Bax} &= MUT_{Bax}*\left(k*\dfrac{p53tt}{k+p53tt}-k*pBax\right)\\
\dot{p Apaf1} &= MUT_{Apaf1}*\left(k*\dfrac{p53tt}{k+p53tt}-k*pApaf1\right)\\
\dot{p CytC} &= k*\dfrac{1}{1+e^{-k*pBax-k}}*k*(1-\dfrac{1}{1+e^{-k*pBcl2-k}})*k*(1-\dfrac{1}{1+e^{-k*pBclXl2-k}})- \\
& kpCytC-k*pApaf1*pCytC^{7}\\
\dot{p Apoptosome} &= k*pApaf1*pCytC^{7}-k*pApoptosome\\
\dot{o Apoptosis} &=k*pFasL+k*pApoptosome-k*oApoptosis\\
\dot{p E2F} &= MUT_{Rb}*MUT_{myc}-kpE2F\\
\dot{p ARF} &= MUT_{arf} \left( k1*\dfrac{pE2F}{k+pE2F}-k2*pARF - k*pARF*pMDM2Nuc \right)\\
\dot{p P21cip} &= k*\dfrac{p53tt}{k+p53tt}-k*pP21cip\\
\dot{p ECDK2} &= k-\dfrac{k*pP21cip}{k+pP21cip}-(k*pECDK2)\\
\dot{p Siah} &= MUT_{Siah}\left(\dfrac{k*p53tt}{k+p53tt}-k*pSiah\right)\\
\dot{p Reprimo} &= MUT_{Reprimo}\left( \dfrac{k*p53tt}{k+p53tt}-k*pReprimo \right)\\
\dot{o Arrestsignal} &= (see~below)\\
\end{align*}

The initial condition of the system is an externally applied radiation dose modeled by setting $oRadiation(0)=1$ followed by an exponential decay. The only other non-zero initial condition is for ECDK2 since at time 0 we assume that there are no brakes on the cell cycle.

\subsubsection{Additional modeling notes}

Cells have many mechanisms to control cell growth and division. We choose to model just a few, and in a simplified manner, to get the flavor of the complexity. We split the control into two cases, one where the Rb gene is functioning ($Rb=1$) and one where the Rb gene is impaired ($Rb=0$). For the $Rb=0$ case, a way to arrest cell growth is via the SIAH or Reprimo gene pathways. SIAH and Reprimo are activated by a functioning p53 danger signal pathway, and we model their effect on the arrest signal as additive. Thus:

Case $Rb=0$:
\begin{equation}
oArrestsignal = \frac{1}{1 + e^{ka1*(x(Siah)+x(Reprimo)-ka2)}};
\end{equation}
We take the derivative of this to embed it into the ODE set.

For $Rb=1$, the Rb controls are working correctly. In that case, low levels of the Cyclin E/CDK2 complex (ECDK2) will arrest the cell cycle, independently of SIAH and Reprimo levels. On the other hand, high levels of ECDK2 mean that the cell can pass through the G1-S transition, but SIAH or Reprimo might still stop it. We model this as a convex combination for the arrest signal:

Case $Rb=1$:
\begin{equation}
  oArrestsignal = \lambda_{low}*1 +(1-
  \lambda_{low}) \frac{1}{1 + e^{ka1*(x(Siah)+x(Reprimo)-ka2)}};
\end{equation}
\noindent where $\lambda_{low} =  1 - (ECDK2/ECDK2_{max})$, where $ECDK2_{max}$ is the maximum level that ECDK2
can attain. We differentiate this as above.

The final classification (into one of four states: 1, 2, 3, or 4) for the ground truth simulation uses the following rules, based on the levels at the end of the simulation:
\begin{verbatim}
If Apoptosis >= 0.8:                    1 (apoptosis)
Else
  If FIXED > 0.9 and ARREST < .5:       2 (repaired and cycling)
  Else if FIXED <= 0.9 and ARREST < .5: 3 (not repaired, and cycling
                                          i.e. mitotic catastrophe)
  Else                                  4 (quiescence)
\end{verbatim}

For details on mutations and parameter changes used for ground truth dataset and the kernel dataset, see the MATLAB input files.

\subsection{Flowering model}
The flowering model is taken directly from \cite{flowering}. The outcome that we build a prediction model
for is flowering time, which, as in the original paper, is taken to be the time at which the protein AP1 exceeds a given threshold. The ODE model is simulated using MATLAB. 
The flowering model represents an  isolated genetic circuit in multi-cellular eukaryote, and therefore as a model is a distant cousin--but a relevant one--to the vastly complicated genetic circuitry of human cancer cells.

\subsection{Boolean cancer model}

The Boolean cancer model is taken from \cite{booleanNetwork}. We converted their GinSIM model into BoolNet format, which is a package in R. The authors provide an original model as well as a modular reduction. In the SimKern simulation we use the modular reduction, which represents a limited understanding of the model, further perturbed by uncertainties of how to map the feature data into this reduced model. The model output for both the ground truth dataset and the SimKern runs are based on the steady state vectors found by simulating the network for the given initial conditions. Let $ss(n)$ denote the steady state value for node $n$. If the steady state is a fixed steady state, $ss(n)$ will be a single value, either 0 or 1. If the steady state is a cycle, then $ss(n)$ will be a binary vector of length equal to the cycle length. For the classification, we rely on two compartments in particular: $n=$~Apoptosis and $n=$~Metastasis. We classify the outputs into three categories using the following logic. 

\begin{verbatim}
If all(ss(Apoptosis)) = 1:        1 (apoptosis)
Else if all(ss(Metastasis)) = 1:  2 (metastasis)
Else                              3 (other)
\end{verbatim}

For details about the meaning of this model we refer the readers to the original publication \cite{booleanNetwork}. In the present work, it is sufficient to view this model as an instance of a discrete complex system.

\subsection{Network flow optimization model}

We wrote a random network generation routine in MATLAB which generates a layer-wise directed graph. The user specifies the number of nodes for each layer and probabilities for adding a connecting arc between the nodes of two layers. We also add arcs between non-adjacent layers with a small probability. We ran this routine once to create a single network for all the samples in the dataset, shown in Figure~S2. We generate random numbers for the cost for these arcs. This represents the base network from which all the samples of SIM0 are built. Unique samples are created by varying the weights of 12 of the 80 arcs, the bold arcs in Figure~S2. The outcome of the simulation is a classification, 1, 2, or 3, representing which of the last three arcs the optimal flow passes through (linear network flow optimization theory guarantees that there exists an optimal solution with all the flow through one of the exit arcs, and that such a solution will be returned by simplex-based methods \cite{lpbook}).

\begin{figure}[!ht]
\centering
\includegraphics[trim=0 0 0 0,clip,width=13cm]{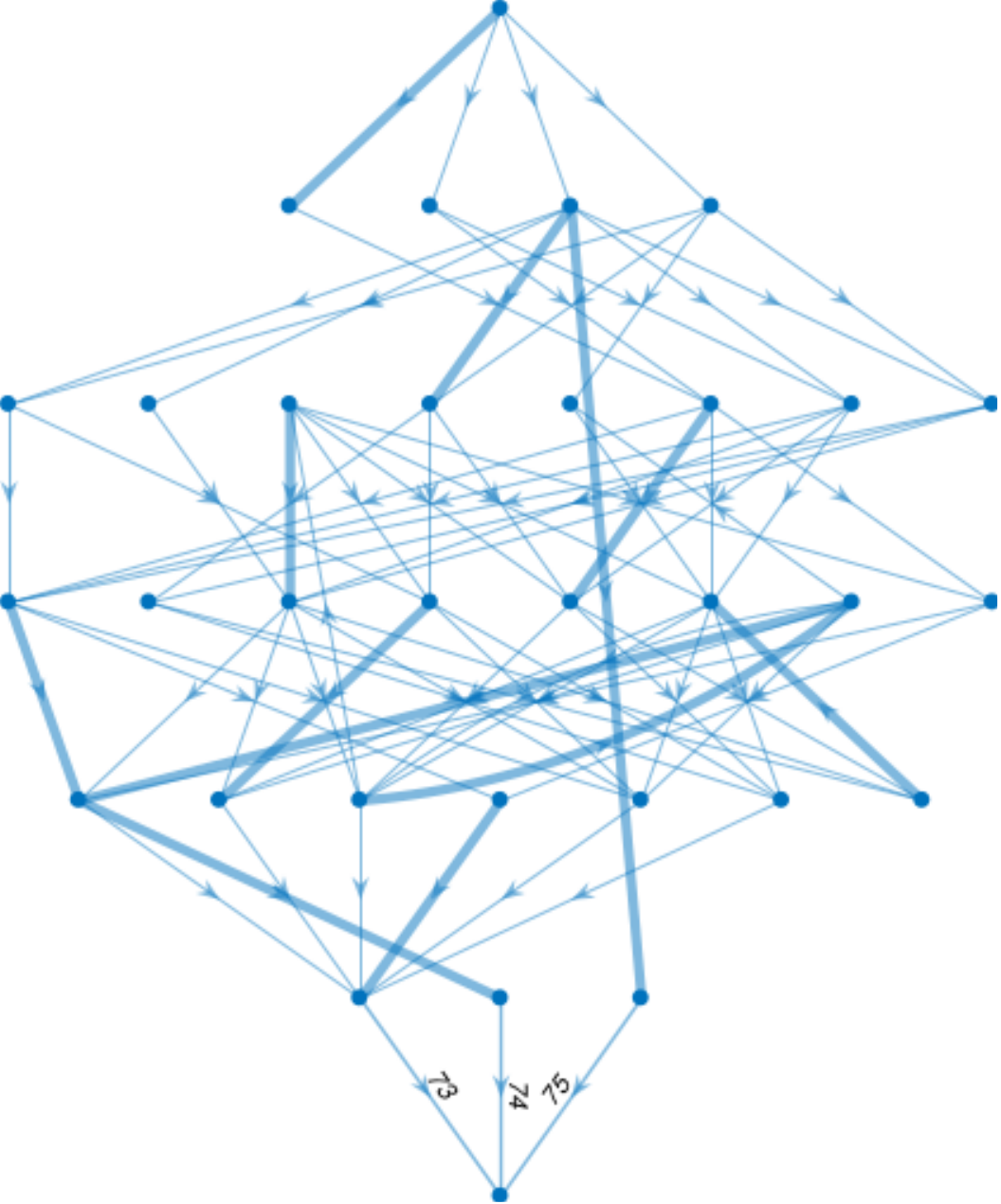}
\label{networkGraph}
\caption{Network flow directed graph. The bold lines are the arcs with variable costs in the ground truth simulation. The unit flow that enters the network at the uppermost node will exit through one of the labeled arcs at the bottom, which creates a classification problem.}
\end{figure}

We run two versions of SIM1, a \emph{less noisy} model (with fewer perturbed, less noisy arc costs) and a \emph{noisier} model (with larger number and higher magnitude of perturbed arc costs). For the \emph{less noisy} models we assume the arc costs of the 12 SIM0 variable arcs are not known with certainty: they are scaled by a uniformly distributed random variable between 0.1 and 1.9. We also perturb every arc in the second layer by a uniform variable from 0.5 to 1.5. For the \emph{noisier} model we additionally perturb the arc costs of the third layer (uniform 0.5 to 1.5) as well as a large perturbation, uniform between 9 and 10, of the third arc, which otherwise always takes the flow because of its otherwise low arc cost (see Figure S4).

\section{Supplementary results}

The flowering model, Figure~S5, displays a typical ``good kernel'' result where the SimKern methods dominate the no-prior-knowledge methods throughout, but especially for small training set sizes. Similarity based NN is competitive with the more sophisticated similarity SVM and RF, but exhibits slightly more variance. The success of the SimKern methods indicates that the space induced by the similarity kernel is well behaved and the classes are easily separable with this kernel.

\begin{figure}[!ht]
\centering
\includegraphics[trim=40 310 60 280,clip,width=16cm]{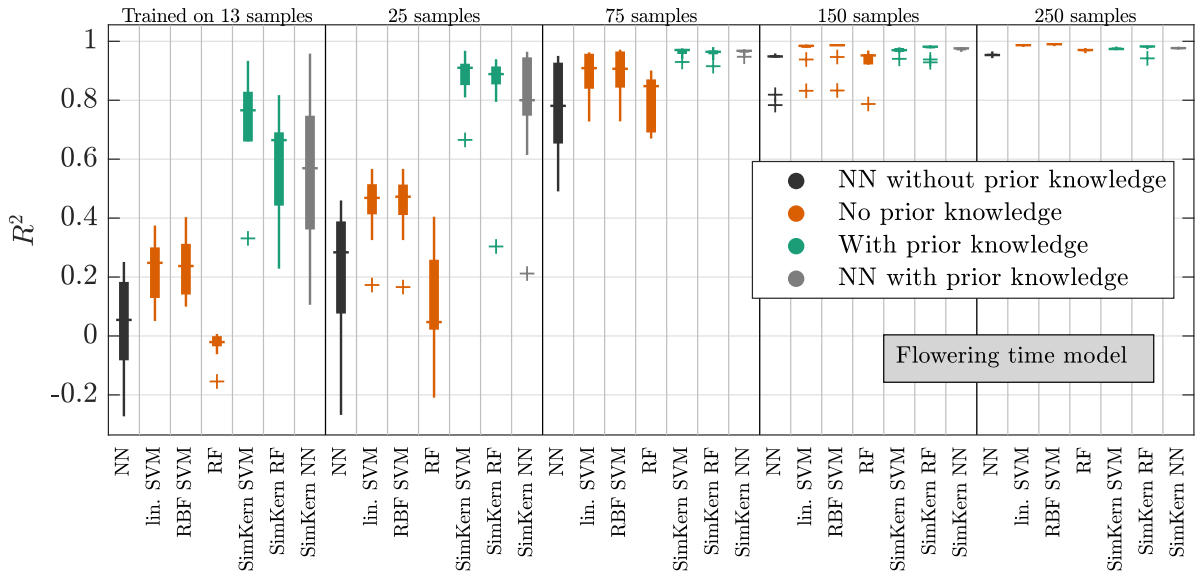} 
\label{floweringBox}
\caption{Machine learning results for the flowering model. NN = nearest neighbor, RF = random forest, SVM = support vector machine, RBF = radial basis function. $R^2$ is the coefficient of determination.}
\end{figure}



With the network flow model, we demonstrate the obvious but important result that if the SimKern simulation is farther from the ground truth simulation due to additional noise, the SimKern learning will be worse. The kernel based on a \emph{less noisy} SimKern simulation, Figure~S6, displays dominance throughout whereas the kernel based on a \emph{noisier} SimKern simulation, Figure~S7, is overtaken by the standard RF already by 18 training samples. We also used vector-based outputs from the SIM1 simulations, where the flow through every arc was used to compute similarity scores. The results were not fundamentally different so here we display results from only the scalar based SIM1 output.

\begin{figure}[!ht]
\centering
\includegraphics[trim=40 310 60 280,clip,width=16cm]{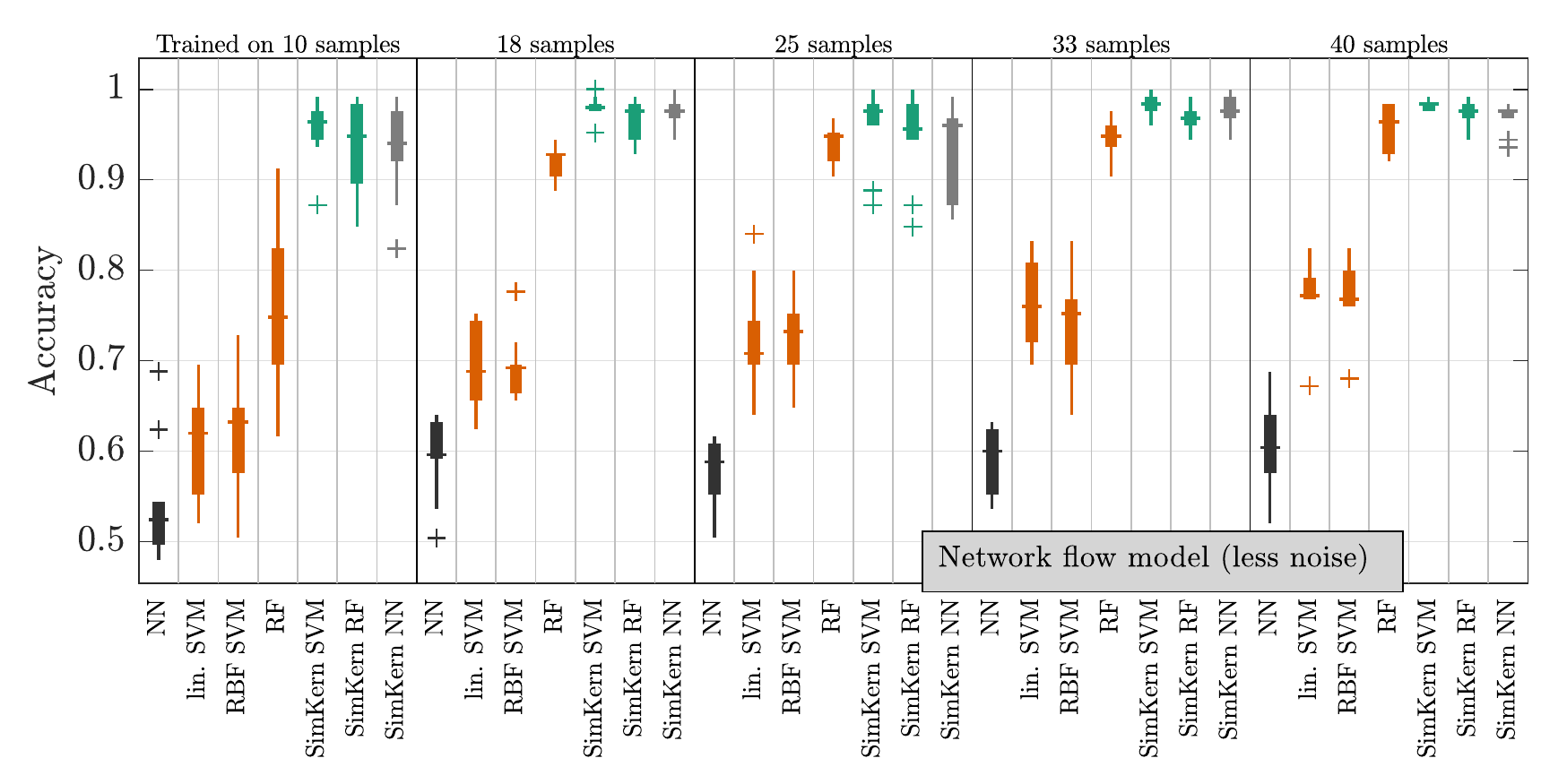} 
\caption{Machine learning results for the network flow optimization model for the \emph{less noise} case. NN = nearest neighbor, RF = random forest, SVM = support vector machine, RBF = radial basis function.}
\label{networkBoxEasier}
\end{figure}

\begin{figure}[!ht]
\centering
\includegraphics[trim=40 310 60 280,clip,width=16cm]{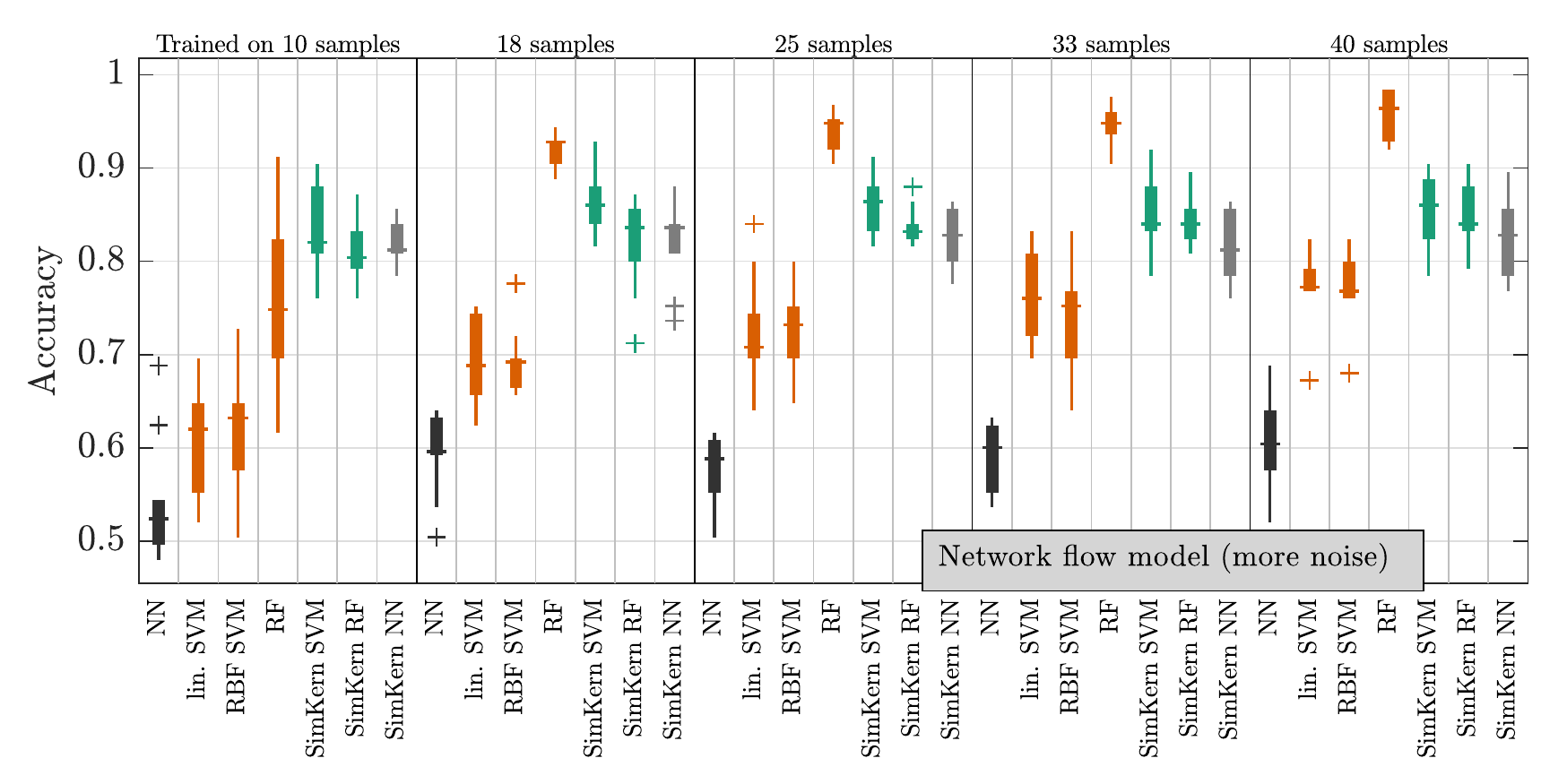} 
\label{networkBoxHarder}
\caption{Machine learning results for the network flow optimization model for the \emph{more noise} case. NN = nearest neighbor, RF = random forest, SVM = support vector machine, RBF = radial basis function. }
\end{figure}


The SimKern idea is effective provided that the simulations correctly judge the similarity between two samples, but the SimKern simulations need not themselves make correct predictions (in fact, the raw output of the SimKern simulations need not be the same type of output as we are trying to predict). To illustrate this, we examine the first 13 samples from the dataset for the network (lower quality) model, see Figure~S8. Samples 2 and 11, which both are classified as 3s in the ground truth dataset, are given a high similarity score because they behave similarly for most of the 10 trials, even though in only one of those trials (trial 6) are they actually classified correctly.

Each model displays two nearest neighbor (NN) algorithm learning results: the default method which is Euclidean distance in feature space, and the kernelized method which uses the simulation based similarity scores for the distance computation. Consistently, the kernel based NN methods dominates over standard NN, which implies the power of a custom similarity measure. The difference between either of these NN methods and the SVMs display the power of better machine learning algorithms: rather than classifying a new sample based on which training sample it is closest to, SVMs factor in the distance to many of the training samples. In some cases (Figure 3, main document: the radiation model with the higher quality kernel, and Figure 4 main document or Figure S5: the flowering model) we see that a good similarity score is ultimately good enough and more advanced machine learning algorithms do not offer much improvement over the kernelized NN. 
 
\begin{figure}[!ht]
\centering
\includegraphics[trim=0 0 0 0,clip,width=13.5cm]{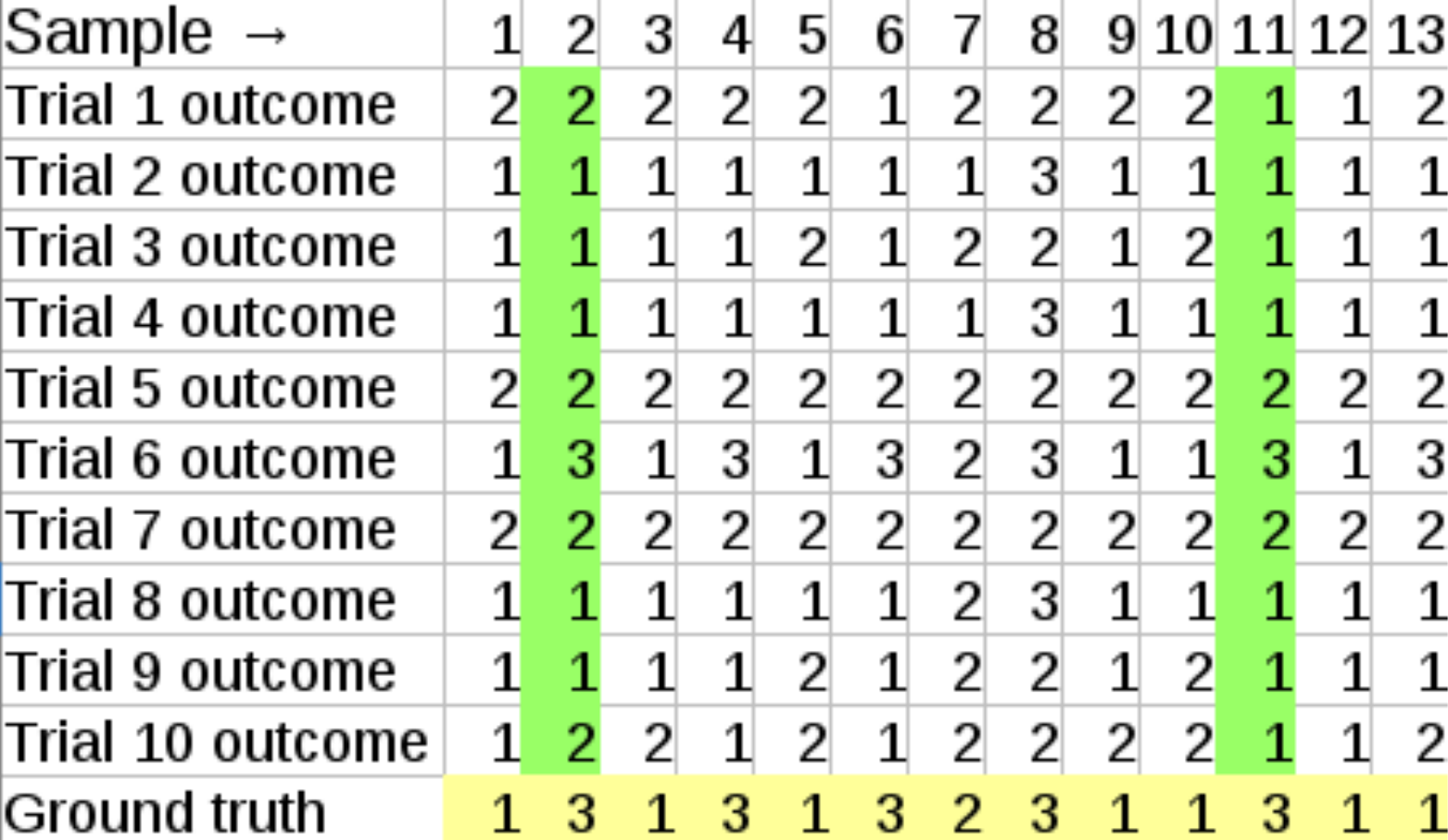}
\label{networkExampleSimilarityExcel}
\caption{SIM1 results for the first 13 samples from the network (lower quality) dataset, for all ten trials and also showing in the bottom yellow row the ground truth (SIM0) result. We have highlighted samples 2 and 11. These samples are both 3s in the ground truth set, but in the $R=10$ SimKern (SIM1) trials they get correctly classified only once. However, they are given a high similarity score since they behave the same for most of the trials. We use this to highlight the idea that it is sufficient to correctly judge sample similarity; accurate class prediction is not necessary.}
\end{figure}

As an additional way to compare machine learning results in the case of regression (the flowering model),
Figure~S9 plots the predicted flowering times versus the actual flowering times.
With additional training samples (13 to 25), linear SVM and, even more so, SimKern SVM improve their predictions for samples with a flowering time $<6$. After training on additional data, one observes a small additional downward bias in linear SVM predictions for samples with a flowering time $>10$. Both algorithms, however, achieve an $R^{2}$ improvement by 0.19 (linear SVM) and 0.31 (SimKern SVM).

\begin{figure}[!ht]
\centering
\includegraphics[trim=60 90 0 80,clip,width=17.5cm]{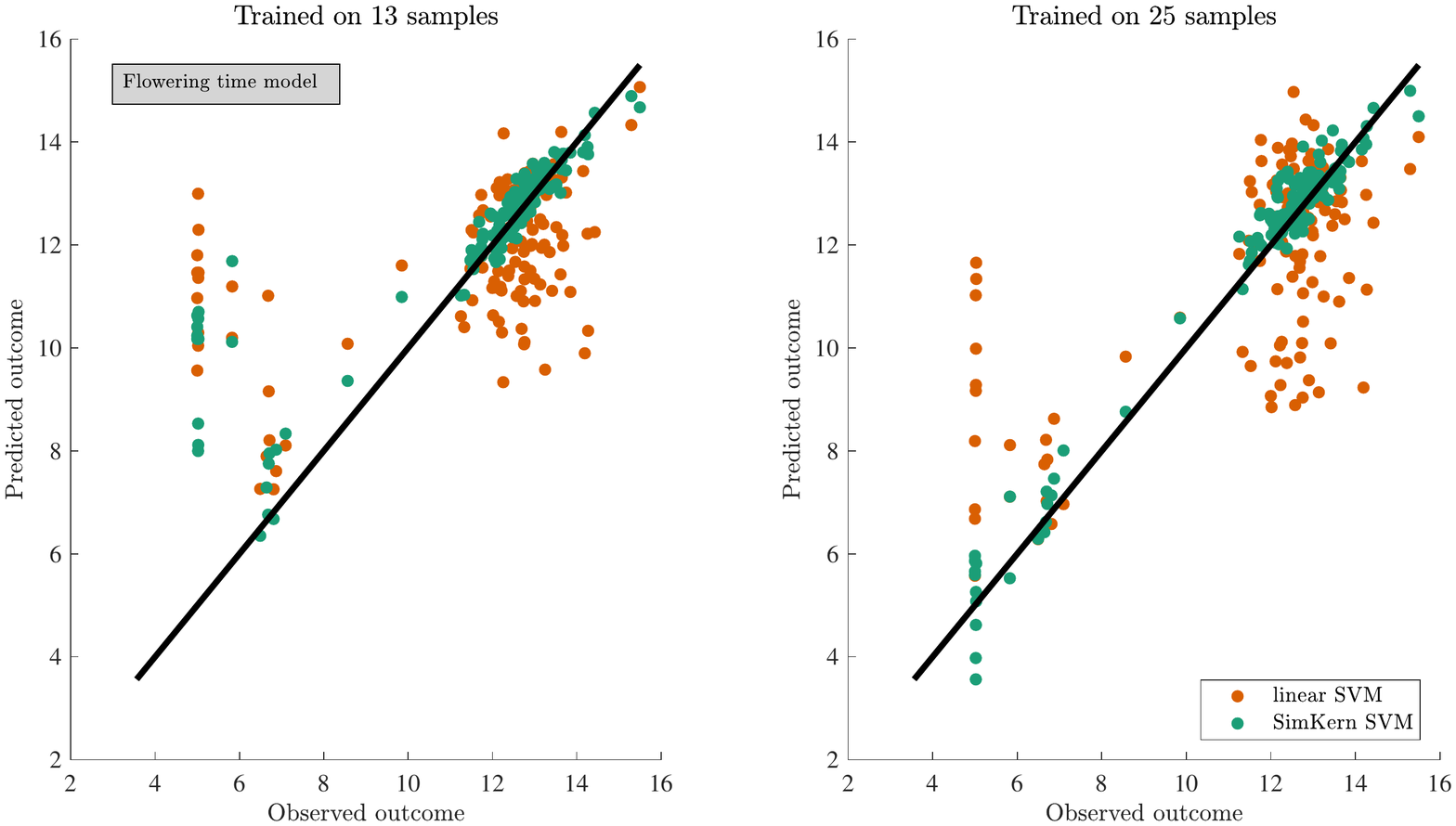}
\label{flowerPredictions}
\caption{Observed and predicted test set values after training on 13 (left) and 25 (right) samples for the flowering time model. Results for linear SVM and SimKern SVM are in orange and green, respectively. Left: $R^{2}$ equals 0.27 and 0.66 for linear SVM and SimKern SVM, respectively. Right: $R^{2}$ equals 0.46 and 0.97 for linear SVM and SimKern SVM, respectively.}
\end{figure}

\section{A word on the ``kernel trick''}

Kernel methods are often touted in the literature as a cure-all for the problem of overly high dimensional samples: by kernelizing the data, the high dimensionality goes away. In fact, kernelizing data does not so cleanly solve this problem since there are many ways to make a kernel. Only when considering highly restricted kernel classes such as linear kernels or RBF kernels, without any feature weighting or feature selection, does the kernel trick simplify the search for a good machine learning approach. But in general, we do not know how to build a good kernel (that is, how to judge similarity between two samples in a way that is most effective for our machine learning problem). We propose herein to distill expert knowledge of a domain into simulations that use the high dimensional features, which pre-supposes quite detailed knowledge of the system. If such detailed knowledge is not available, the number of ways to turn a large feature set into a kernel is unmanageable (consider combinatoric calculations for example of selecting 200 genes out of 20000 to test all sets of 200 genes). We state this here as a word of caution: the kernel approach can be very useful but it requires obtaining a good kernel, and there is no general recipe for that.

Clearly for the SimKern approach to work, the simulations used to generate the kernel have to be ``good'', but unfortunately, it does not seem possible to be more quantitative than that for general cases. We explored the issue by demonstrating that as we veer away from high quality simulations, the machine learning using the custom kernel does worse (see e.g. Figures ~S6 and S7), but it will always be a data- and problem-specific analysis to see if a proposed kernel is useful.

\bibliography{allpics}
\bibliographystyle{unsrt}